\documentclass[10pt,twocolumn,letterpaper]{article}

\usepackage[pagenumbers]{cvpr} 

\newcommand{\hide}[1]{}
\usepackage{multirow}
\usepackage{makecell}

\definecolor{cvprblue}{rgb}{0.21,0.49,0.74}
\usepackage[pagebackref,breaklinks,colorlinks,allcolors=cvprblue]{hyperref}
\usepackage{svg}
\usepackage{colortbl}

\def\paperID{11674} 
\def\confName{CVPR}
\def\confYear{2026}

\title{SPOT: Sparsification with Attention Dynamics via Token Relevance in\\Vision Transformers}

\author{
Oded Schlesinger$^{1,2}$ \quad
Amirhossein Farzam$^{1,2}$ \quad
J. Matias Di Martino$^{1,3}$ \quad
Guillermo Sapiro$^{2,4}$
\\[2mm]
$^{1}$Duke University \quad
$^{2}$Princeton University \quad
$^{3}$Universidad Católica del Uruguay \quad
$^{4}$Apple
}

\begin{document}
\maketitle
\begin{abstract}
While Vision Transformers (ViT) have demonstrated remarkable performance across diverse tasks, their computational demands are substantial, scaling quadratically with the number of processed tokens.
Compact attention representations, reflecting token interaction distributions, can guide early detection and reduction of less salient tokens prior to attention computation\hide{, leading to significant efficiency gains}.
Motivated by this, we present \textbf{SP}arsification with attenti\textbf{O}n dynamics via \textbf{T}oken relevance (\mbox{SPOT}), a framework for early detection of redundant tokens within ViTs that leverages token embeddings, interactions, and attention dynamics across layers to infer token importance, resulting in a more context-aware and interpretable relevance detection process.
\mbox{SPOT} informs token sparsification and facilitates the elimination of such tokens, improving computational efficiency without sacrificing performance.
\mbox{SPOT} employs computationally lightweight predictors that can be plugged into various ViT architectures and learn to derive effective input-specific token prioritization across layers.
Its versatile design supports a range of performance levels adaptable to varying resource constraints.
Empirical evaluations demonstrate significant efficiency gains of up to 40\% compared to standard ViTs, while maintaining or even improving accuracy.
Code and models are available at~\url{https://github.com/odedsc/SPOT}\hide{https://anonymous.4open.science/r/SPOT}.
\end{abstract}

\section{Introduction}\label{sec:introduction}
\begin{figure}[t!]
    \centering
    \begin{tabular}{>{\centering\arraybackslash}m{0.9\linewidth} >{\centering\arraybackslash}m{0.12\linewidth}}
        \hspace{-0.45cm}\includegraphics[width=\linewidth]{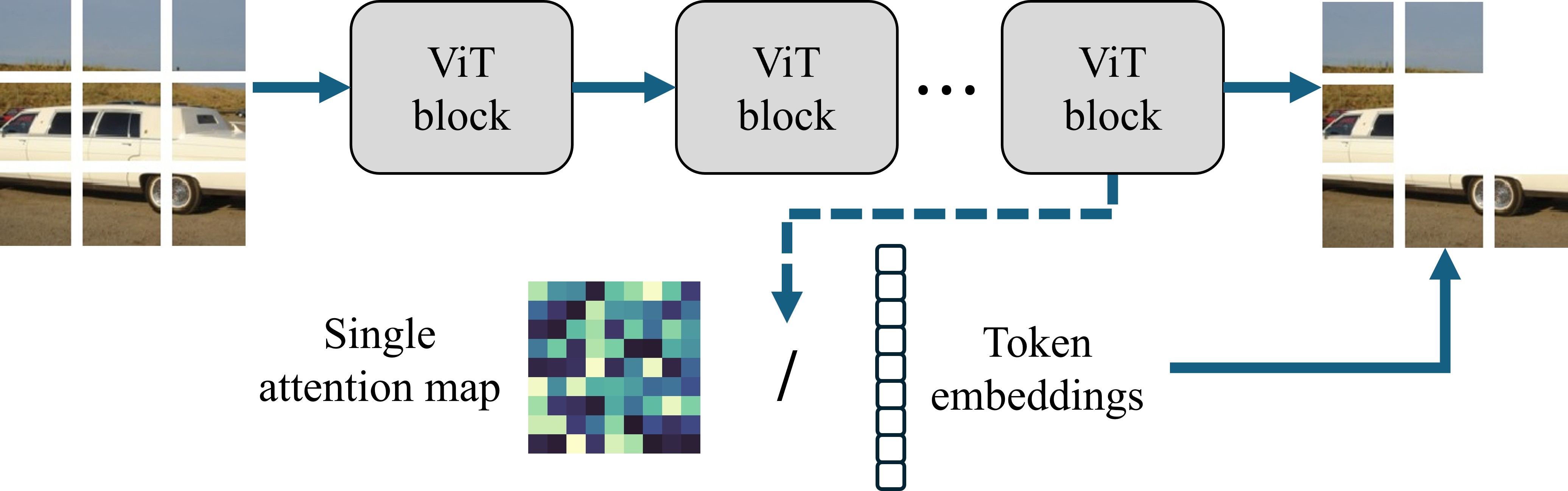} &
        \hspace{-1.3cm}\includegraphics[width=\linewidth]{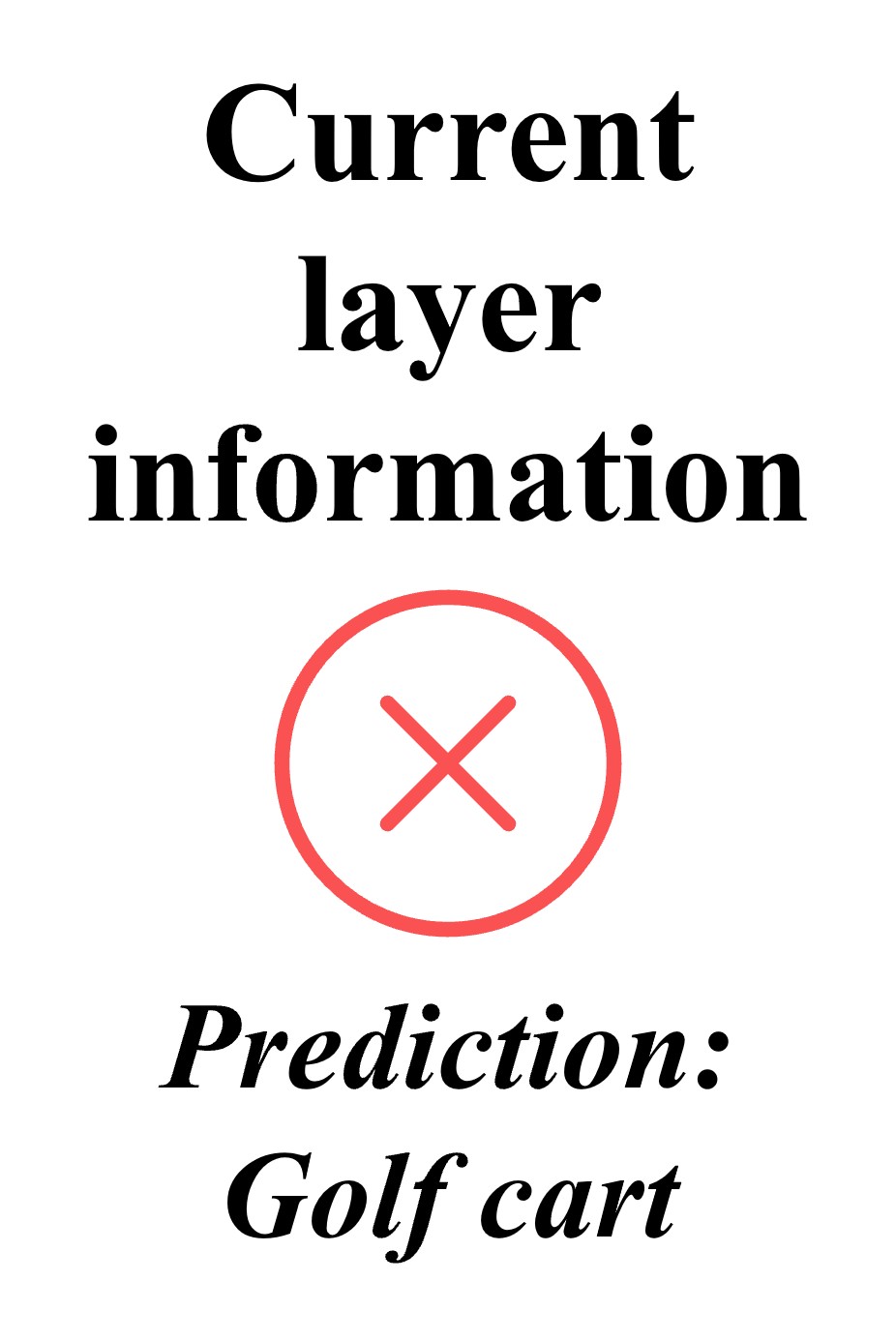} \\
    \end{tabular}
    
    \vspace{-0.02cm}
    \includegraphics[width=1.005\linewidth]{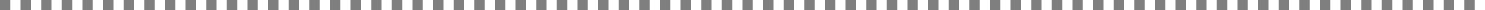}\vspace{0.1cm}
    
    \begin{tabular}{>{\centering\arraybackslash}m{0.9\linewidth} >{\centering\arraybackslash}m{0.12\linewidth}}
        \hspace{-0.45cm}\includegraphics[width=\linewidth]{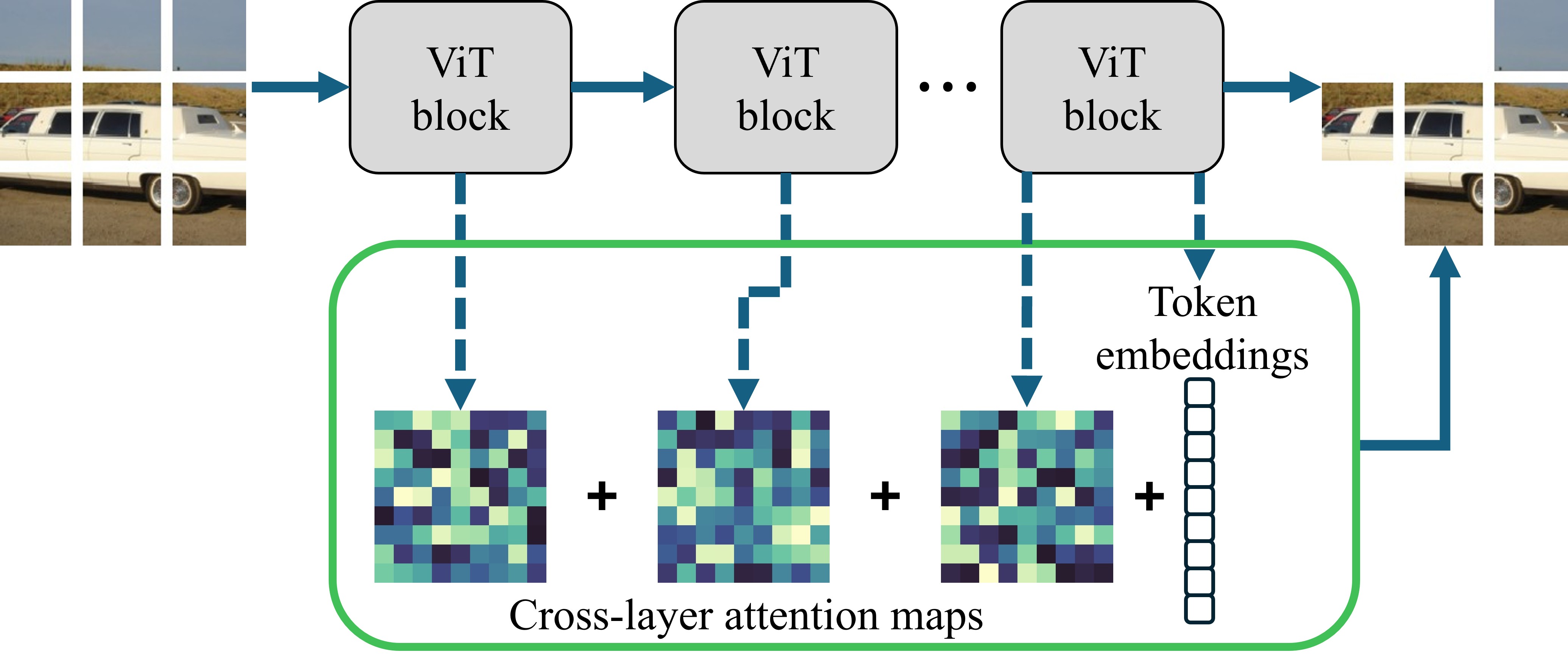} &
        \hspace{-1.3cm}\includegraphics[width=\linewidth]{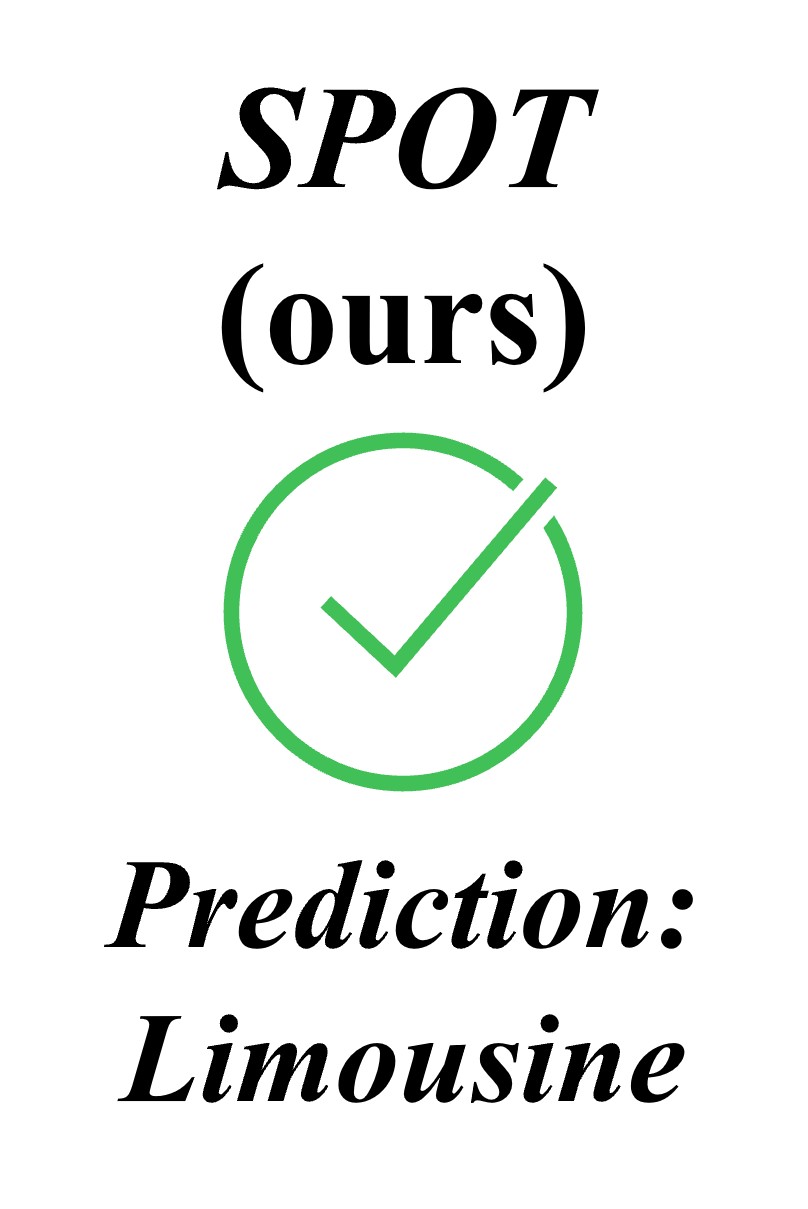}
    \end{tabular}
    
    \vspace{-0.1cm}
    \includegraphics[width=1.005\linewidth]{figures/1/separator.jpg}\vspace{0.1cm}
    \begin{tabular}{lr}
         \hspace{-0.225cm}\includegraphics[width=0.495\linewidth]{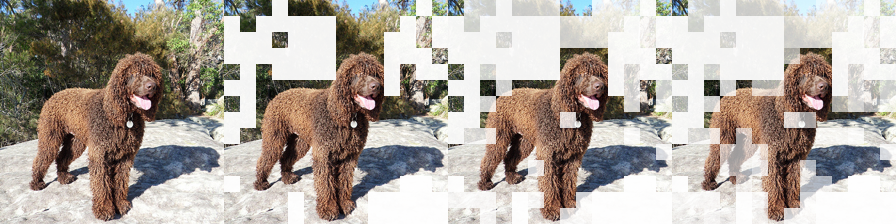} &  
         \hspace{-0.325cm}\includegraphics[width=0.495\linewidth]{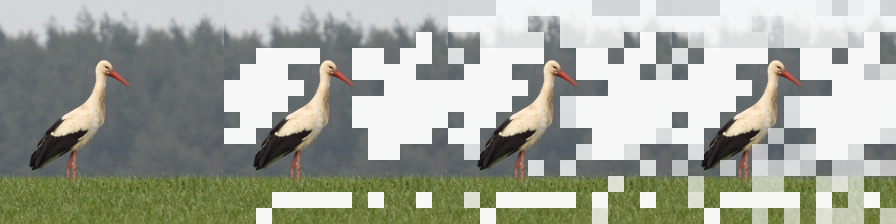}
    \end{tabular}
    \vspace{-0.3cm}
    \caption{Overview of the proposed \mbox{\textit{SPOT}} framework.
    Top: current sparsification methods primarily rely on partial information from the current model layer—either token embeddings or attention maps—to determine token importance, potentially leading to suboptimal predictions.
    Middle: \mbox{\textit{SPOT}} enhances this process by integrating compact representations of both token embeddings and attention maps and their dynamics from multiple ViT blocks (highlighted in green) to \mbox{\textit{SPOT}} and select the most crucial input-specific information. Providing this multi-layer information facilitates the gradual pruning of less relevant tokens, improving efficiency while maintaining accuracy.
    \hide{The limo example image was partitioned into 9 tokens for visualization purposes.\\}
    Bottom: Example visualizations of the gradual relevance-driven token selection performed by \mbox{\textit{SPOT}}.}
    \label{fig:introduction}
\end{figure}

Transformers~\cite{vaswani2017attention} have become a leading neural network architecture, demonstrating outstanding performance across diverse tasks. This success is largely attributed to their powerful attention mechanism, which allows dynamic weighting of input interactions.
While Convolutional Neural Networks (CNNs) have proven effective in various computer vision applications~\cite{he2016deep,krizhevsky2012imagenet,lecun1998gradient,simonyan2014very}, their inherent locality limits global context modeling.
In contrast, the attention mechanism facilitates simultaneous consideration of all input elements, effectively modeling long-range dependencies.
This ability has proven crucial for understanding complex visual scenes.
Following the success of Vision Transformers (ViT)~\cite{dosovitskiy2021an}, which partition each image into a set of patches forming a sequence of tokens, transformer-based models have achieved state of the art performance across various tasks~\cite{bao2023all, carion2020end, esser2021taming, jiang2021all, liu2021swin, ramesh2022hierarchical, strudel2021segmenter, touvron2021training, touvron2021ICCV, xie2021segformer, zhang2022dino, zheng2021rethinking, zhu2020deformable}.

However, this strength comes at a significant computational cost.
The quadratic complexity of attention with respect to the number of input tokens necessitates substantial computational resources~\cite{shen2021efficient}, limiting their scalability and posing a significant bottleneck in resource-constrained settings.
This computational burden stems from the indiscriminate processing of all input image regions, including those that are less informative to the task at hand.
A further challenge in utilizing attention mechanisms within ViTs stems from the presence of spikes with seemingly spurious high-attention values that do not correlate with semantically relevant image features~\cite{darcet2023vision,guo2023robustifying,yang2025denoising}.
This phenomenon can hinder the utilization of attention maps, particularly when relying on a single model state for decision-making.

This paper addresses the computational bottleneck of ViTs by introducing an effective method for the early detection of less informative tokens.
Our token relevance prediction mechanism enables efficient and robust salient token selection based on contextual and dynamics information across network layers, as illustrated in Figure~\ref{fig:introduction}. 
We demonstrate that this relevance prediction can inform sparsification during inference, maintaining or even improving model accuracy with considerable computational savings.

\begin{figure*}[t]
    \centering
    \includegraphics[width=0.985\textwidth]{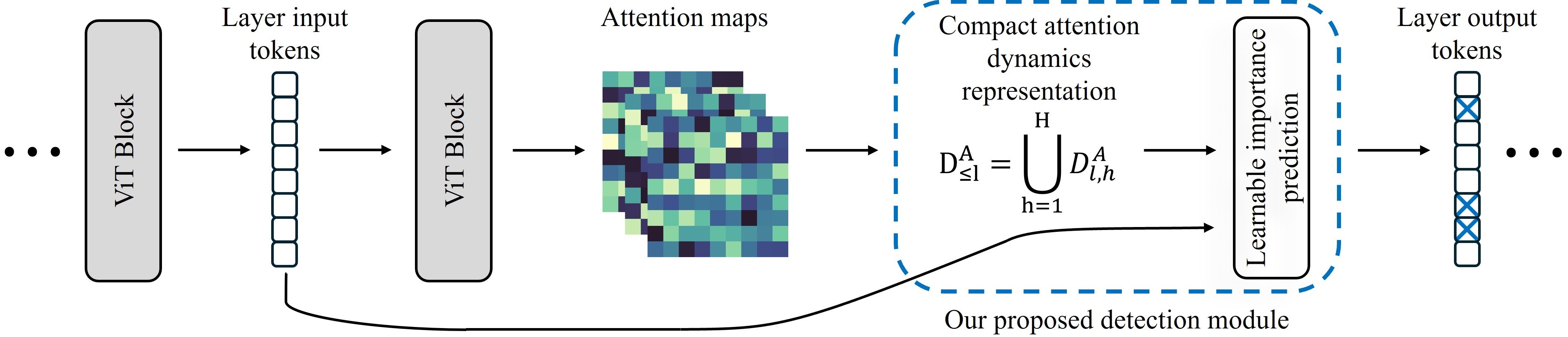}
    \vspace{-0.1cm}
    \caption{Illustration of the proposed redundant token detection data flow.
    Our modular design enables effective redundant tokens identification within ViT-based architectures by plugging the proposed module (blue) into any of the model's self-attention layers (gray). 
    At each stage, the method leverages a combination of token embeddings and attention map statistics across layers until the current one from each head, as described in Section~\ref{subsec:method}.
    This provides the \mbox{\textit{SPOT}} module with rich contextual information and a comprehensive representation of inter-token interactions and their dynamics throughout the model, enabling it to predict redundant tokens effectively.}
    \label{fig:pipeline}
\end{figure*}

Our proposed \textbf{SP}arsification with attenti\textbf{O}n dynamics via \textbf{T}oken relevance (\mbox{\textit{SPOT}}) framework leverages token-derived information and pre-existing attention maps without altering the underlying ViT backbone to detect less informative tokens early in the attention pipeline.
To accomplish this, attention values are aggregated across layers, yielding an effective representation of feature importance and mitigating the effects of attention inter-layer noise fluctuations (see Appendix~\ref{supp:sec:theoretical} for details).
Specifically, \mbox{\textit{SPOT}} is equipped with information regarding the dynamic evolution of attention distributions across model layers.
This approach exploits inter-layer attention dynamics, effectively capturing how the relative importance of tokens changes throughout the processing pipeline, as illustrated in Figure~\ref{fig:pipeline}.

We revisit the attention mechanism application within the transformer architecture, maximizing the utilization of all pre-computed attention values from preceding layers.
Instead of discarding these computationally expensive intermediate representations after a single processing step, \mbox{\textit{SPOT}} further exploits this information for enhanced performance.
To this end, \mbox{\textit{SPOT}} incorporates the moments of the distribution of the attention received and originating from the tokens, considering each attention map in the current and preceding layers.
By maintaining and addressing attention dynamics throughout the transformer blocks, \mbox{\textit{SPOT}} informs token relevance prediction with a more holistic perspective.
This, in turn, leads to better-informed hierarchical token sparsification and prevents the premature removal of tokens due to fluctuations in importance across layers.

Our main contributions are:
\begin{itemize}
    \item We introduce \textit{\mbox{\textit{SPOT}}}, a framework for detecting the importance of individual input tokens while modeling cross-layer attention dynamics and leveraging comprehensive token-level and attention-derived information.
    \item We present a modular, computationally inexpensive token prioritization module that can be plugged into existing ViT architectures, enabling substantial computational savings without requiring structural modifications.
    \item Comprehensive experimental results demonstrate the efficacy and versatility of our proposed method in detecting less informative tokens in ViT models, aligning with human-interpretable feature relevance and facilitating efficient deployment in resource-constrained applications.
\end{itemize}

\section{Related Work}\label{sec:related_work}
\noindent\textbf{Vision Transformers.}
Transformers have revolutionized computer vision, delivering state-of-the-art performance across various tasks. 
Their adoption began with image classification, where ViTs~\cite{dosovitskiy2021an} demonstrated competitive results compared to CNNs--the dominant architecture prior to transformers.
Subsequent research, e.g., Swin Transformer~\cite{liu2021swin} and CaiT~\cite{touvron2021ICCV}, introduced hierarchical structures and improved training stability.
Transformers have also shown exceptional promise in object detection~\cite{carion2020end, zhang2022dino, zhu2020deformable}, semantic segmentation~\cite{caron2021emerging,strudel2021segmenter,xie2021segformer, zheng2021rethinking}, and image generation~\cite{ramesh2022hierarchical, esser2021taming}.
This diverse body of work underscores the broad applicability of transformers across computer vision domains.
However, the high computational cost of the attention mechanism remains a significant barrier.

\noindent\textbf{Efficient ViTs.}
Different techniques have been explored to mitigate the computational demands of ViTs and improve model efficiency.
DeiT~\cite{touvron2021training} introduced efficient training strategies aiming to reduce the reliance on extensive computational resources and data, while LV-ViT~\cite{jiang2021all} investigated token labeling strategies and employed convolutions.
Other works approximated the attention mechanism using low-rank matrices~\cite{wang2020linformer}, locality-sensitive hashing~\cite{kitaev2020reformer}, or alternative formulations~\cite{choromanskirethinking,shen2021efficient}.
Complementary computational optimization techniques such as caching~\cite{gemodel}, tiling~\cite{dao2022flashattention}, and chunking~\cite{kitaevreformer} further enhance computational and memory efficiency.
Another notable approach is reducing the number of computed attention interactions through sparsification.
Early methods, e.g.,~\citet{child2019generating}, employed fixed, heuristic-based sparsity patterns\hide{, such as restricting attention to neighboring tokens only}, limiting their adaptability.
Subsequent research has shifted toward detecting redundant tokens dynamically, allowing input-dependent sparsity patterns, often referred to as pruning, by processing and adjusting token sets based on input features and network state.

\noindent\textbf{Hard vs. Soft Sparsification.}
\textit{Hard sparsification} is a category of these methods which removes identified redundant tokens~\cite{rao2021dynamicvit, pan2021ia, yin2022vit, liang2022not, wang2024zero, liu2024a, lu2025reinforcement}, preventing their further processing in subsequent attention blocks.
Conversely, \textit{soft sparsification} methods take further steps to generate new tokens from these tokens, typically through modifications to the attention mechanism, including token compression~\cite{wu2023ppt, chen2023diffrate}, fusion~\cite{kim2024token, wei2023joint}, packaging~\cite{kong2022spvit}, merging~\cite{bolya2022token}, or aggregation~\cite{xu2022evo}.
Both categories are fundamentally dependent on the accurate and early identification of redundant tokens.
While soft sparsification methods introduce post-identification processing, they often rely on the same relevance predictions as hard sparsification methods.

This work addresses the critical challenge of redundant token detection by offering a modular foundation that can also be integrated to enhance soft sparsification.
Focusing on this task facilitates a direct and more interpretable demonstration of computational efficiency improvements via a masking strategy employed in hard sparsification, as visualized in Figure~\ref{fig:visualization}.
Nevertheless, to comprehensively evaluate the efficacy of the proposed relevance prediction method, we conducted experiments using both hard and soft sparsification techniques, thereby demonstrating a versatile token prioritization mechanism alongside complementary benefits for refined optimization in existing methods.

\noindent\textbf{Token Redundancy Detection.}
Existing detection methods differ in their primary information source, typically utilizing either token-level features or attention-based relationships.
Within the category of token-based approaches, DynamicViT~\cite{rao2021dynamicvit} employs simple prediction modules using current layer tokens, adding low computational overhead.
A-ViT~\cite{yin2022vit} proposes computing halting probabilities for each token using their embeddings.
RL4EViT~\cite{lu2025reinforcement} incorporates reinforcement learning to facilitate token pruning decisions.
Methods relying solely on token embeddings may overlook the broader dynamics of tokens across the network, which is, in part, the core of the attention mechanism.
Attention-driven strategies, exemplified by EViT~\cite{liang2022not}, typically assess token importance based on the class token attention scores. This is often followed by preserving the top-$K$ tokens exhibiting the highest attention values.
Evo-ViT~\cite{xu2022evo} offers residual connections between class token attention scores across layers to maintain the information flow within the network.
Kong et al.~\cite{kong2022spvit} enable a more nuanced consideration by considering tokens' relevance to each attention head.
PPT~\cite{wu2023ppt} also considers the values matrix, $V$, and proposes a more comprehensive token scoring metric.
Nevertheless, these approaches overlook the information embedded within the full attention maps.

Other sparsification methods typically base token importance decisions on these token and attention-derived foundations, e.g., dTPS~\cite{wei2023joint} utilizes DynamicViT~\cite{rao2021dynamicvit} predictors.
In contrast, our proposed \mbox{\textit{SPOT}} leverages broader input dynamics and unifies both token embeddings and the rich information from all attention maps, within and across layers, in a compact fashion.
This hybrid strategy allows us to analyze inter-token dynamics and efficiently identify tokens that contribute minimally to the task at hand.

\section{Methods}\label{sec:methods}
\subsection{Preliminaries}\label{subsec:preliminaries}
ViT~\cite{dosovitskiy2021an} employs a tokenization strategy that divides images into non-overlapping patches, linearly projecting each patch into a latent token embedding.
A learnable class token, [cls], is prepended to this sequence to facilitate global image representation.
These tokens are processed by the model, a core element of which is the self-attention mechanism, which quantifies inter-token relationships.
The self-attention mechanism first projects the input tokens via learnable projection matrices to obtain the query, $Q$, and key, $K$ matrices, and then computes their similarity scaled by the square root of their dimensionality, $d_k$, followed by a softmax normalization:
\begin{equation}\label{eq:attn_map}
    A(Q, K) = softmax\left(\frac{QK^T}{\sqrt{d_k}}\right).
\end{equation}
This operation, performed across all token pairs within each layer and head of the architecture, yields a set of square attention matrices, enabling the model to capture complex, long-range dependencies within the image.
The final output of the attention module is then obtained by multiplying $A(Q, K)$ with a value matrix $V$:
\begin{equation}\label{eq:attn_output}
    Attn(Q, K, V) = A(Q, K) \cdot V.
\end{equation}

\subsection{Proposed Methodology}\label{subsec:method}
We offer a modular approach that does not modify the backbone of the attention model through which we sparsify the input tokens.
Our proposed lightweight module provides information derived from attention maps by leveraging and re-utilizing information already computed within the model, which is typically discarded after a single forward pass.
To predict what tokens would be the most crucial in subsequent model layers for a given task, we introduce a hybrid approach considering both token embedding representations and the attention map in the current layer, along with those from preceding layers.
These attention dynamics through the layers of the transformer serve as a historical context for how attention is propagated.
To this end, we aggregate this information via the statistical moments of attention score distributions across the current and preceding layers.

To formalize this process, let $N_l$ be the number of input patch tokens in layer $l$, and $A_{l,h}$ the $(N_l+1) \times (N_l+1)$ attention map for head $h$.
Following~\citet{caron2021emerging}, this matrix is partitioned into three relevant sub-regions, reflecting distinct attention relationships: $A_{l,h}^{cls, out}$, capturing the attention weights from the class token to all other tokens; $A_{l,h}^{cls, in}$, capturing the attention weights from all other tokens to the class token; and $A_{l,h}^{non\_cls}$, the submatrix representing interactions between all non-class tokens.
These can be written over the attention map as:
\begin{equation}
    A_{l,h}^{cls, out} = A_{l,h}[1, 2:N_l+1],
\end{equation}
\begin{equation}
    A_{l,h}^{cls, in} = A_{l,h}[2:N_l+1, 1],
\end{equation}
where we exclude the self-interaction term of the class token, represented by the top-left element of $A_{l,h}$.

We incorporate dynamics awareness by extracting aggregated information from the attention maps (see Figure~\ref{fig:pipeline}).
For each head, we first concatenate the first row and first column vectors of these matrices, representing the class token interactions with the image patch tokens.
\begin{figure*}[t]
    \centering
    \begin{tabular}{cc}
        \vspace{-0.05cm}
        \includegraphics[width=0.46\linewidth]{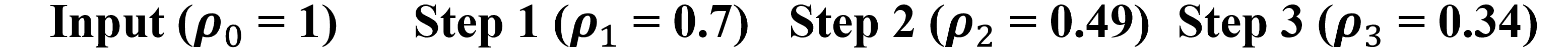} & \hspace{-0.4cm} \includegraphics[width=0.46\linewidth]{figures/visualization_S/Text2.png} \\
        \vspace{-0.05cm}
        \includegraphics[width=0.46\linewidth]{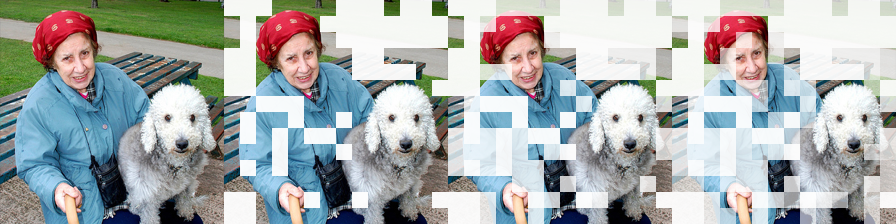} & \hspace{-0.4cm} \includegraphics[width=0.46\linewidth]{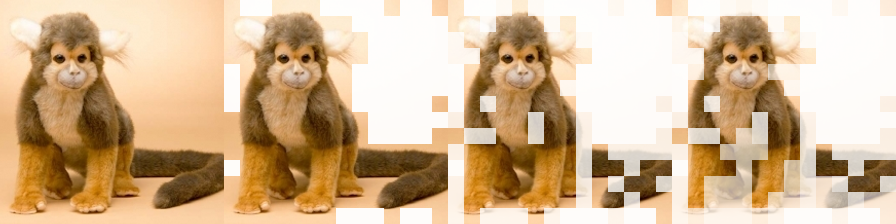} \\
        \vspace{-0.05cm}
        \includegraphics[width=0.46\linewidth]{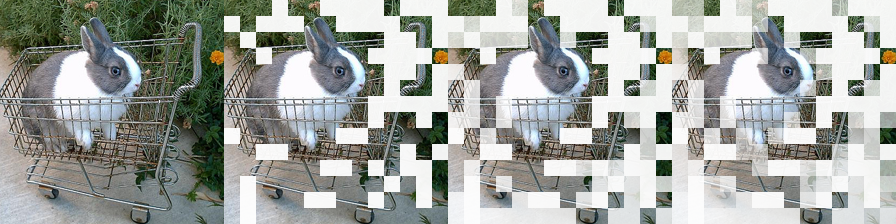} & \hspace{-0.4cm} \includegraphics[width=0.46\linewidth]{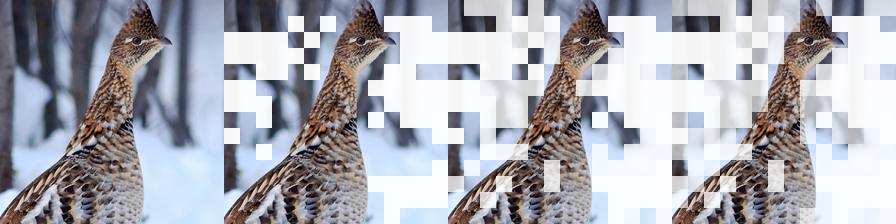} \\
        \vspace{-0.05cm}
        \includegraphics[width=0.46\linewidth]{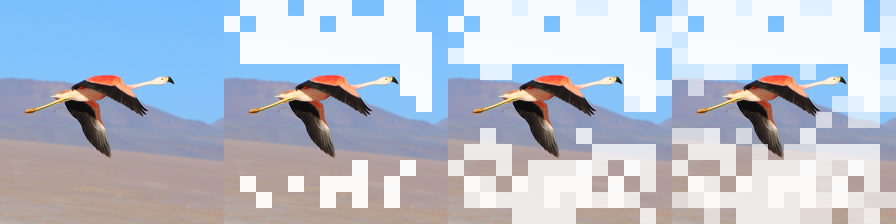} & \hspace{-0.4cm} \includegraphics[width=0.46\linewidth]{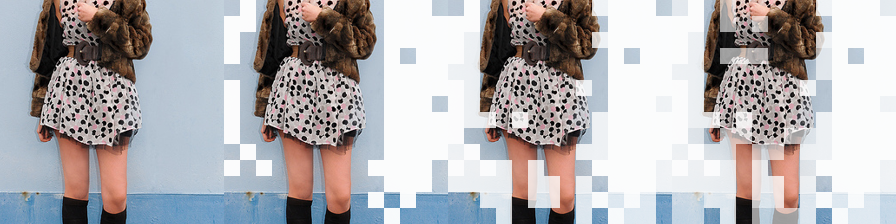} \\
        \vspace{-0.05cm}
        \includegraphics[width=0.46\linewidth]{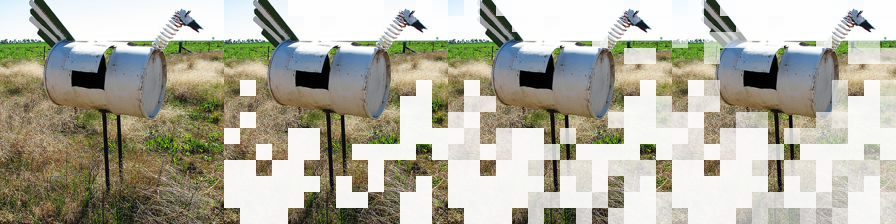} & \hspace{-0.4cm} \includegraphics[width=0.46\linewidth]{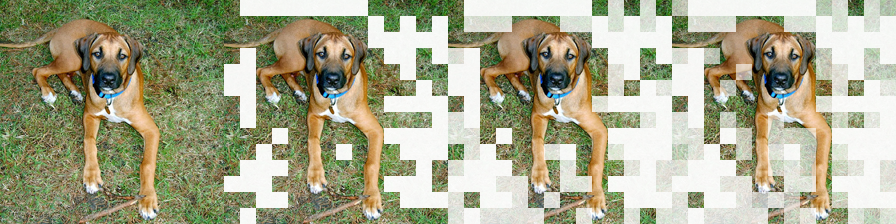} \\
        \vspace{-0.05cm}
        \includegraphics[width=0.46\linewidth]{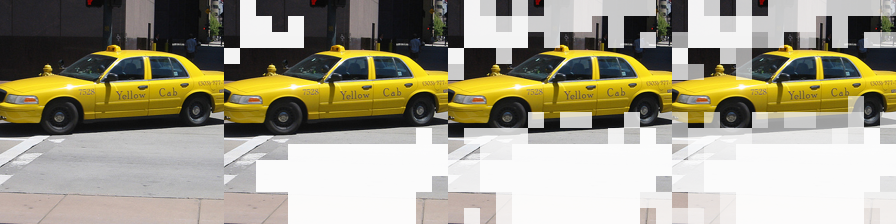} & \hspace{-0.4cm} \includegraphics[width=0.46\linewidth]{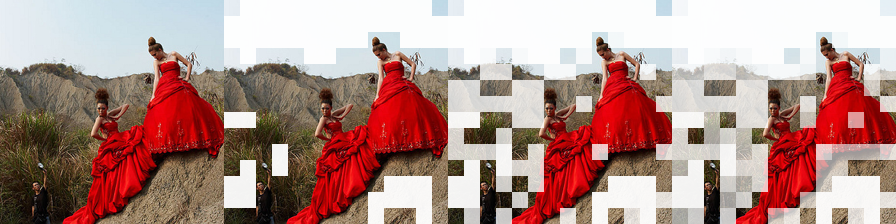} \\
    \end{tabular}
    \vspace{-0.15cm}
    \caption{Visualizations of the gradual redundant token detection performed by our proposed approach on DeiT-S on samples from ImageNet-1K validation set.
    Increasingly transparent masking shades indicate later detection.
    Tokens identified as more informative, and thereby retained, are well aligned with semantic image objects and visual features, pointing to \mbox{\textit{SPOT}}'s interpretability.}
    \label{fig:visualization}
\end{figure*}
The remaining elements of each attention map, $A_{l,h}^{non\_cls}$, are then summarized by computing row-wise and column-wise mean and variance.
The corresponding statistics for each token $t$ are:
\begin{equation}\label{eq:mu_row}
    {\mu}_{l,h}^{A, row}[t] = \frac{1}{N_l} \cdot \sum_{j=1}^{N_l}{A_{l,h}^{non\_cls}[t,j]},
\end{equation}
\begin{equation}\label{eq:mu_col}
    {\mu}_{l,h}^{A, col}[t] = \frac{1}{N_l} \cdot \sum_{i=1}^{N_l}{A_{l,h}^{non\_cls}[i,t]},
\end{equation}
\begin{equation}\label{eq:sigma_row}
    {\sigma}_{l,h}^{A, row}[t] = \sqrt{\frac{1}{N_l} \cdot \sum_{j=1}^{N_l}{{(A_{l,h}^{non\_cls}[t,j]-{\mu}_{l,h}^{A, row}[t])}^2}},
\end{equation}
\begin{equation}\label{eq:sigma_col}
    {\sigma}_{l,h}^{A, col}[t] = \sqrt{\frac{1}{N_l} \cdot \sum_{i=1}^{N_l}{{(A_{l,h}^{non\_cls}[i,t]- {\mu}_{l,h}^{A, col}[t])}^2}}.
\end{equation}
Concatenating these scalar token-wise statistics for all $N_l$ tokens yields four vectors, ${\mu}_{l,h}^{A, row},
{\mu}_{l,h}^{A, col},
{{\sigma}_{l,h}^{A, row}},
{{\sigma}_{l,h}^{A, col}}$.
By incorporating the first two moments of the attention distribution, these vectors effectively summarize both incoming and outgoing attention patterns of each token, providing a compact representation of the underlying interaction dynamics within the current layer.
The extracted data from $A_{l,h}$ is then represented as an $N_l \times 6$ concatenation of vectors of the form
\begin{equation}
    D_{l,h}^A = [
    A_{l,h}^{cls, out},
    A_{l,h}^{cls, in},
    {\mu}_{l,h}^{A, row},
    {\mu}_{l,h}^{A, col},
    {{\sigma}_{l,h}^{A, row}}^2,
    {{\sigma}_{l,h}^{A, col}}^2
    ].
\end{equation}

The mean and variance matrices are constructed by calculating the corresponding quantities from the input layer up to (and including) layer $l$ over retained tokens. These matrices, denoted by $M_{l,h}$ and $\Sigma_{l,h}$, respectively, are partitioned and processed following the procedure described above, yielding $D_{l,h}^M$ and $D_{l,h}^\Sigma$.
This characterization of prior attention distributions aims to capture the dynamics and interaction-aware contextual understanding of inter-token relationships across the network layers.

Input features also include remapped token representations, which were projected via a fully connected layer into a $d_\text{remap}$-dimensional space.
Each remapped token embedding is partitioned into global and local features, denoted as $z_\text{global}$ and $z_\text{local}$, respectively, both with $\frac{d_{\text{remap}}}{2}$ dimensionality, consistent with the methodology outlined in DynamicViT~\cite{rao2021dynamicvit}.
Concatenated with token embeddings, we obtain the resulting per-head feature vector of layer $l$ encapsulating the token characteristics, token interactions, and the attention dynamics, as described by
\begin{equation}\label{eq:predictor_info}
    D_{l,h} = [
    z_\text{global},
    z_\text{local},
    D_{l,h}^A,
    D_{l,h}^M,
    D_{l,h}^\Sigma
    ].
\end{equation}
This holistic and efficient summary compactly equips the module with rich, context-aware information by integrating token-specific features with attention within and across transformer blocks, thereby facilitating informed redundant token detection without incurring significant computational overhead. The resulting vector is then processed by a multilayer perceptron (MLP), which serves as a learned importance prediction function.
The architectural specifications of the MLP are detailed in Appendix~\ref{supp:subsec:architecture}, supplemented by a theoretical perspective provided in Appendix~\ref{supp:sec:theoretical}.

The sparsification step entails filtering tokens that the above-described relevance prediction finds to be less informative across multiple layers.
This process is characterized by a retention rate, $\rho$, which governs the proportion of tokens considered more informative and retained at each decision detection iteration.
Specifically, at iteration $k$, the effective retention rate is $\rho_k = \rho ^ k$.
Subsequently, a binary mask, $B_k \in \{0, 1\}^{N_l}$, is derived for each of the training samples, resulting in $|S|$ masks across the training set for each detection iteration.
These masks indicate whether each token is retained or removed during sparsification, except for the class token, which is always retained.
To maintain differentiability during training, $B_k$ is obtained using Gumbel-Softmax sampling~\cite{jang2016categorical} applied to the output of the MLP.
The masks are hierarchically compatible, such that if a token is masked at iteration $k$, it remains masked in all subsequent iterations $T>k$.
Therefore, filtered tokens are subsequently excluded from further processing, concentrating computational resources on more salient token subsets.

The resulting training regime involves optimizing the relevance prediction modules and adapting the model backbone to accommodate token reduction through fine-tuning.
This fine-tuning process minimizes an objective function (Equation~\ref{eq:objective_function}) comprising several components: a task loss (Equation~\ref{eq:classification_loss}), a sparsity rate deviation loss (Equation~\ref{eq:deviation_loss}), and loss terms promoting similarity to an unaltered pre-trained model (equations~\ref{eq:prediction_similarity_loss} and~\ref{eq:token_similarity_loss}).
For a classification task, given the predicted classes $y$, and ground truth labels $y_{true}$, a typical task loss is the cross-entropy
\begin{equation}\label{eq:classification_loss}
    L_{cls} = CrossEntropy(y_{true}, y).
\end{equation}

\begin{figure*}[t]
    \centering
    \begin{tabular}{cc}
        \vspace{-0.15cm}
        \includegraphics[height=3.1cm]{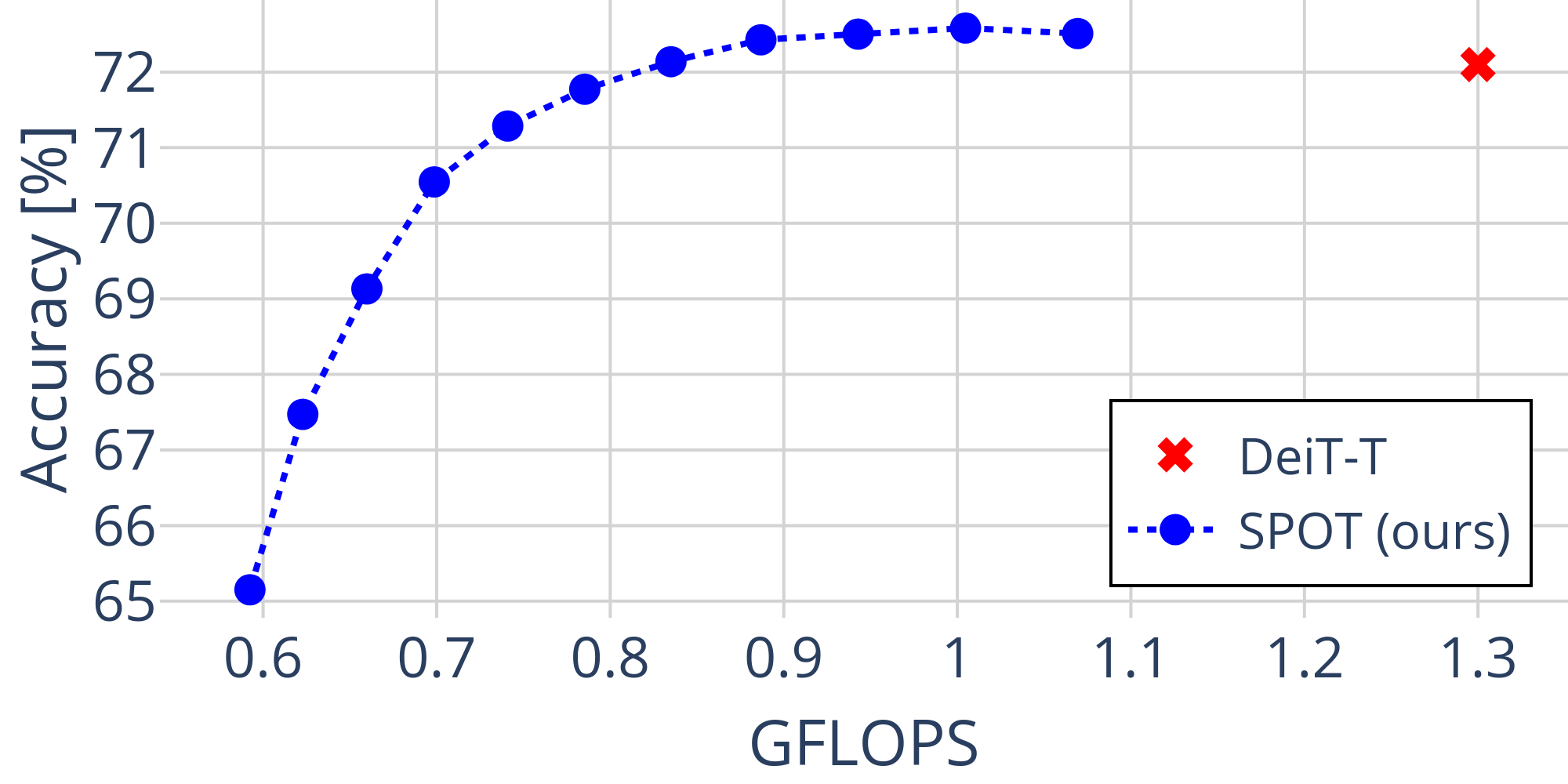} & \hspace{0.5cm} \includegraphics[height=3.1cm]{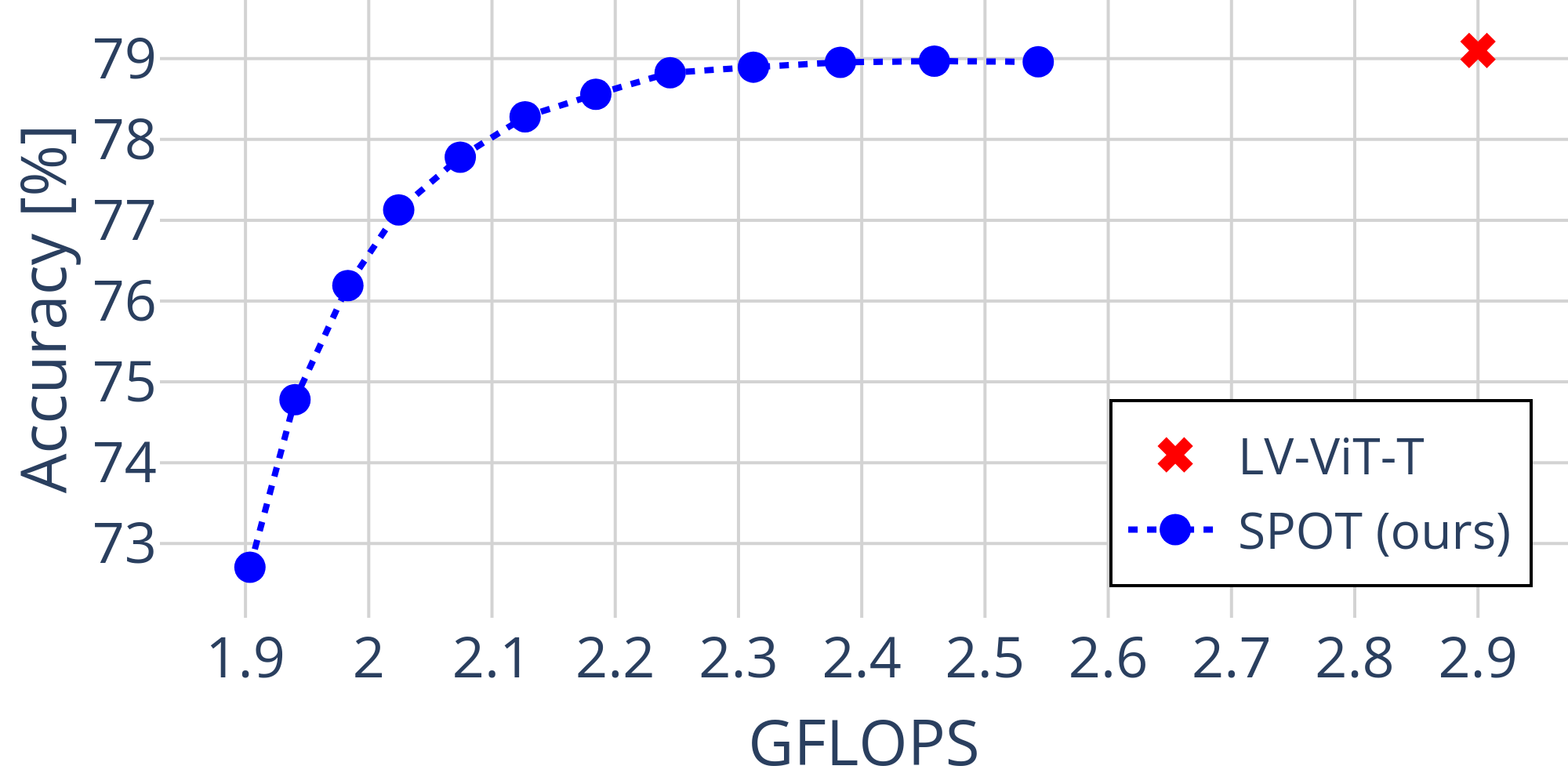} \\
        \vspace{-0.125cm}
        \includegraphics[height=3.1cm]{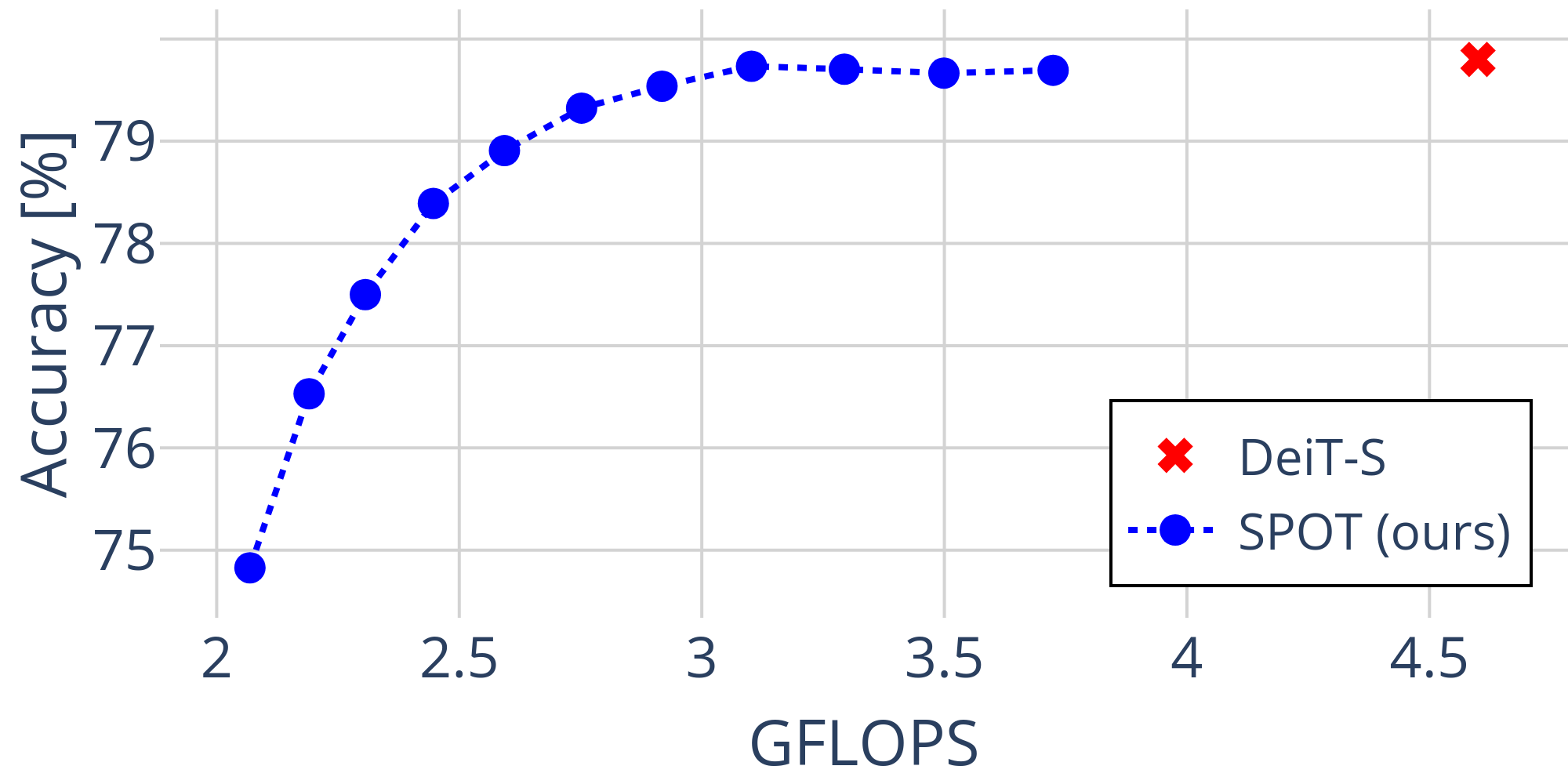} & \hspace{0.5cm} \includegraphics[height=3.1cm]{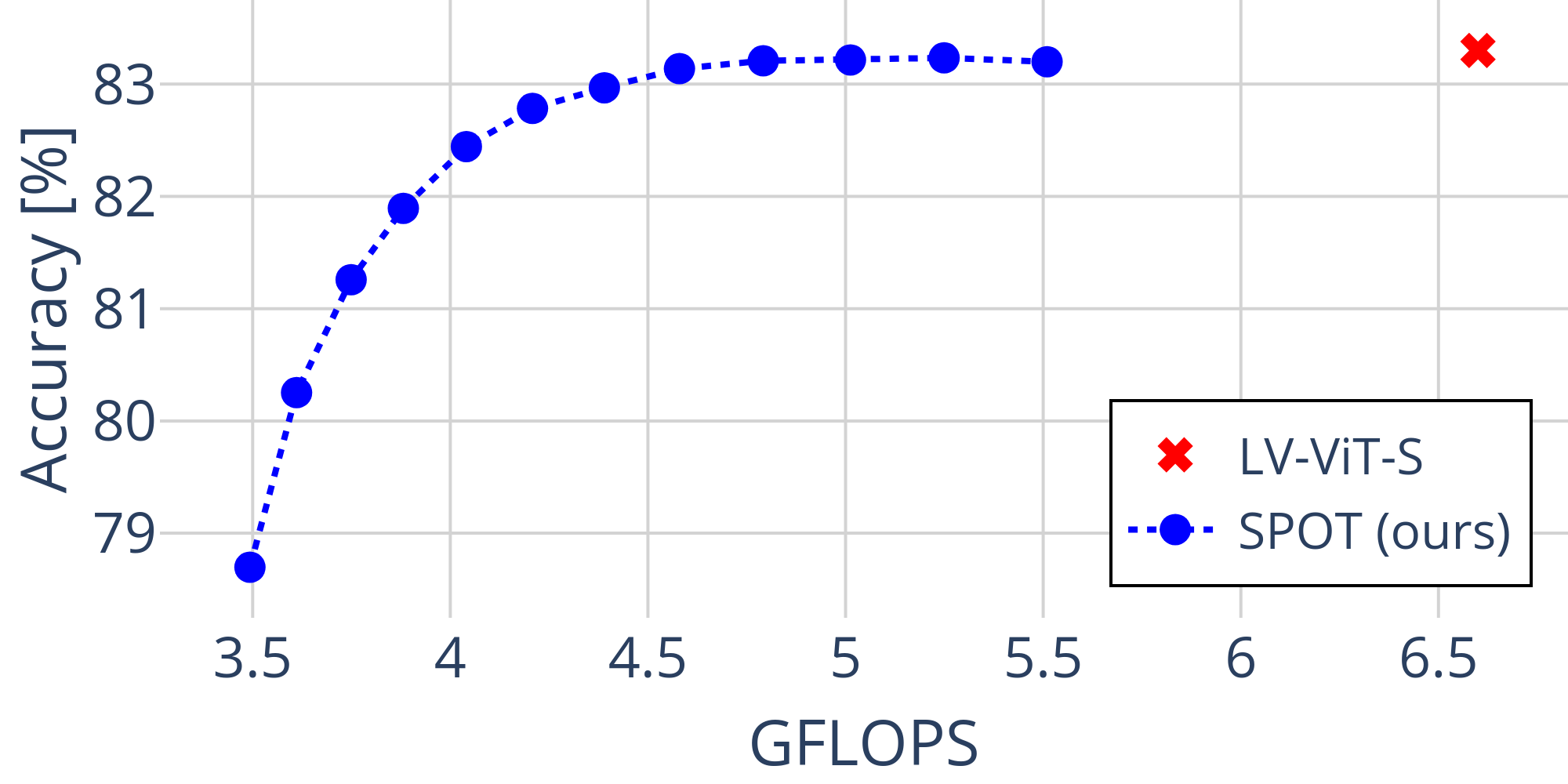} \\
    \end{tabular}
    \vspace{-0.15cm}
    \caption{Performance of \mbox{\textit{SPOT}} on the ImageNet-1K dataset.
    We evaluate \mbox{\textit{SPOT}} classification accuracy across four different models: DeiT-T (top-left), DeiT-S (bottom-left)~\cite{touvron2021training}, LV-ViT-T (top-right), and LV-ViT-S (bottom-right)~\cite{jiang2021all}, under varying computational budgets quantified in GFLOPS, corresponding to different retention rates, set by $\rho$.
    As expected, our framework exhibits a trade-off between efficiency and performance, as higher computational budgets lead to higher accuracy.
    }
    \label{fig:main_results}
    \vspace{-.15cm}
\end{figure*}

The sparsity rate deviation loss regulates the discrepancy between the empirical token retention rate $\{\hat{\rho}_k\}_{k=1}^{K}$ derived from the input specific predicted mask $\{B_k\}_{k=1}^{K}$, and target retention rates, $\{{\rho}_k\}_{k=1}^{K}$, across $K$ sparsification iterations and $|S|$ training samples. This can be expressed as
\begin{equation}\label{eq:deviation_loss}
    L_{rate} = \frac{1}{|S| \cdot K} \cdot \sum_{m=1}^{|S|} \cdot \sum_{k=1}^{K} {(\rho_{k,m} - \hat{\rho}_{k,m}})^2.
\end{equation}
The prediction similarity loss is the Kullback–Leibler divergence between the pre-trained model's prediction probabilities $y'$, and those of the sparsified model $y$
\begin{equation}\label{eq:prediction_similarity_loss}
    L_{pred} = D_{KL}(y'||y).
\end{equation}
Finally, the token similarity term is given by the cosine similarity in last-layer token representations between the unaltered model's tokens $t'_m$ and the fine-tuned model's tokens $t_m$, considering only retained tokens,
\begin{equation}\label{eq:token_similarity_loss}
    L_{token} = \frac{1}{|S|} \cdot \sum_{m=1}^{|S|} \left(1-\frac{t'_m \cdot t_m}{||t'_m|| \cdot ||t_m||}\right).
\end{equation}
The total loss is a weighted sum of these terms, given by
\begin{equation}\label{eq:objective_function}
    L = L_{cls} + \lambda_1 L_{rate} + \lambda_2 L_{pred} + \lambda_3 L_{token},
\end{equation}
where the values $\lambda_1,\ \lambda_2$, and $\lambda_3$ were empirically set.

\section{Experiments}\label{sec:experiments}
Our main experiments compare \mbox{\textit{SPOT}} with existing sparsification methods, alongside qualitative observations on \mbox{\textit{SPOT}}'s performance, and ablation studies including experiments evaluating \mbox{\textit{SPOT}} and its versatility on various benchmark datasets.
Additional experiments highlighting \mbox{\textit{SPOT}}’s efficacy and robustness, including evaluations of accuracy under various image perturbations, comparison of different attention and dynamics aggregation strategies, smaller-scale model redundant token identification visualizations, and a throughput analysis, are provided in Appendix~\ref{supp:sec:additional}.
Together, these findings provide comprehensive empirical evidence supporting the effectiveness and generalizability of \mbox{\textit{SPOT}} across multiple ViT architectures and datasets.

\begin{table}[b]
    \resizebox{\linewidth}{!}{
    \begin{tabular}{l|lll}
    Model & Method & Accuracy (\%) & GFLOPS \\
    
    \hline
    \multirow{6}{*}{DeiT-T}
     & Baseline~\cite{touvron2021training} & 72.2 & 1.3 \\
     & DynamicViT*~\cite{rao2021dynamicvit} & 71.4 (-0.8) & 0.8 (-0.5) \\
     & A-ViT~\cite{yin2022vit} & 71.0 (-1.2) & 0.8 (-0.5) \\
     & EViT*~\cite{liang2022not} & 71.9 (-0.3) & 0.8 (-0.5) \\
     & Zero-TPrune~\cite{wang2024zero} & 70.4 (-1.8) & 0.9 (-0.4) \\
     & \textbf{\mbox{\textit{SPOT}} (ours)} & \textbf{72.3 (+0.1)} & 0.8 (-0.5) \\
     
    \hline
    \multirow{10}{*}{DeiT-S}
     & Baseline~\cite{touvron2021training} & 79.8 & 4.6 \\
     & DynamicViT~\cite{rao2021dynamicvit} & 79.3 (-0.5) & 2.9 (-1.7) \\
     & A-ViT~\cite{yin2022vit} & 78.6 (-1.2) & 3.6 (-1.0) \\
     & EViT~\cite{liang2022not} & 79.4 (-0.4) & 3.0 (-1.6) \\
     & $\text{IA-RED}^2$~\cite{pan2021ia} & 79.1 (-0.7) & 3.2 (-1.4) \\
     & RL4EViT~\cite{lu2025reinforcement} & 79.2 (-0.6) & 3.0 (-1.6) \\
     & Zero-TPrune~\cite{wang2024zero} & 79.4 (-0.4) & 3.1 (-1.5) \\
     & METR~\cite{liu2024a} & 79.5 (-0.3) & 3.0 (-1.6) \\
     & \textbf{\mbox{\textit{SPOT}} (ours)} & 79.5 (-0.3) & \textbf{2.8 (-1.8)} \\
     & \textbf{\mbox{\textit{SPOT}} (ours)} & \textbf{79.7 (-0.1)} & 3.0 (-1.6) \\

    \hline
    \multirow{3}{*}{LV-ViT-T}
     & Baseline~\cite{jiang2021all} & 79.1 & 2.9 \\
     & DynamicViT*~\cite{rao2021dynamicvit} & 77.1 (-2.0) & 2.0 (-0.9) \\
     & \textbf{\mbox{\textit{SPOT}} (ours)} & \textbf{77.6 (-1.5)} & 2.0 (-0.9) \\

    \hline
    \multirow{4}{*}{LV-ViT-S}
     & Baseline~\cite{jiang2021all} & 83.3 & 6.6 \\
     & DynamicViT~\cite{rao2021dynamicvit} & 83.0 (-0.3) & 4.6 (-2.0) \\
     & EViT~\cite{liang2022not} & 83.0 (-0.3) & 4.7 (-1.9) \\
     & \textbf{\mbox{\textit{SPOT}} (ours)} & \textbf{83.1 (-0.2)} & \textbf{4.5 (-2.1)} \\
    \end{tabular}
    }
    \vspace{-0.15cm}
    \caption{Results of hard sparsification methods on DeiT and LV-ViT models over ImageNet-1K.
    Comparison is conducted between methods with comparable computational constraints.
    \mbox{\textit{SPOT}} achieves superior results\hide{ among hard sparsification methods}, yielding a notable and consistent increase in accuracy alongside larger computational savings.
    Starred results (*) were achieved via reimplementation using official code.}
    \label{table:comparison_hard}
\end{table}

\subsection{Experimental Setup}\label{subsec:setup}
We conduct image classification experiments on the ImageNet-1K dataset~\cite{deng2009imagenet}, unless otherwise stated, using two ViT-based architectures, DeiT~\cite{touvron2021training} and LV-ViT~\cite{jiang2021all}, with their tiny and small-sized variations, denoted by T and S, respectively.
Models were fine-tuned following standard protocols; full training and optimization details are provided in Appendix~\ref{supp:subsec:implementation}.
Following standard practice, model performance is assessed using classification accuracy, while computational complexity is quantified by giga floating-point operations per second (GFLOPS) during inference.

All models described herein were initialized from pre-trained models.
Subsequently, both the prediction modules and the transformer model underwent joint optimization and fine-tuning with a retention rate of $\rho = 0.7$, over $K=3$ sparsification iterations.

\subsection{Results}\label{subsec:results}
Figure~\ref{fig:visualization} presents a qualitative visualization of our token identification process applied to samples from the ImageNet-1K dataset.
This visualization showcases that our proposed method progressively eliminates image patches that correspond to regions of lower semantic saliency and importance, highlighting its interpretability.

Figure~\ref{fig:main_results} presents the performance of the \mbox{\textit{SPOT}} framework on DeiT and LV-ViT model architectures.
Our evaluation was conducted across a range of $\rho$ values, enabling comprehensive performance analysis under varying computational budgets, balancing accuracy and GFLOPS.
As empirically shown in Figure~\ref{fig:main_results}, our proposed method facilitates significant efficiency enhancements, substantially reducing computational load while concurrently preserving high levels of model accuracy.
A comparative analysis across architectures and scales, presented in Table~\ref{table:comparison_hard}, reveals that our proposed \mbox{\textit{SPOT}} framework exhibits notable improvements in computational efficiency while achieving superior performance compared to state-of-the-art hard token sparsification techniques.
Specifically, \mbox{\textit{SPOT}} attains up to a $40\%$ reduction in GFLOPS (e.g., from 4.6 to 2.8 in DeiT-S).
As demonstrated in Figure~\ref{fig:main_results}, this computational gain can be further increased when adjusting the efficiency and performance trade-off.
In addition to efficiency and accuracy, \mbox{\textit{SPOT}} exhibits consistency and robustness to perturbations (see Table~\ref{table:perturbations}), maintaining performance on noisy and corrupted inputs, and interpretability across model scales.

Importantly, the proposed predictor introduces only minimal internal overhead of 0.032, 0.051, 0.129, and 0.129 GFLOPS for \mbox{DeiT-T}, \mbox{LV-ViT-T}, \mbox{DeiT-S}, and \mbox{LV-ViT-S}, respectively.
These amount to approximately $4\%$ of the models’ GFLOPS and parameter counts.
Such a negligible cost is effectively outweighed by the substantial computational savings predictors achieve through token sparsification, as shown in Table~\ref{table:comparison_hard}.

\begin{table}[b]
    \centering
    \resizebox{0.8\linewidth}{!}{%
    \begin{tabular}{l|lll}
    \begin{tabular}[c]{@{}l@{}}Removed\\ from $D$\end{tabular} & \begin{tabular}[c]{@{}l@{}}Accuracy\\ (\%)\end{tabular}& \begin{tabular}[c]{@{}l@{}}Predictor\\ GFLOPS\end{tabular} & \begin{tabular}[c]{@{}l@{}}Overall\\ GFLOPS\end{tabular} \\ \hline
    $\hat{\mu}$ & 79.4 (-0.4) & 0.120 & 3.0 (-1.6) \\
    $\hat{\sigma}^2$ & 79.5 (-0.3) & 0.120 & 3.0 (-1.6) \\
    $D^M, D^V$ & 79.5 (-0.3) & 0.111 & 3.0 (-1.6)
    \end{tabular}
    }
    \caption{Reduced attention information results demonstrating the efficacy of our compact attention dynamics representation.
    This representation alone already provides substantial information to the token importance predictors, resulting in computational gains.}
    \label{table:attention_ablation_study}
    \vspace{-0.15cm}
\end{table}

Notably, when applied to DeiT-T, \mbox{\textit{SPOT}} leads to simultaneous accuracy and efficiency gains
This likely stems from an implicit regularization effect; by pruning and eliminating less informative tokens that may act as noise, \mbox{\textit{SPOT}} reduces the model's tendency to overfit and enhances generalization, a benefit particularly pronounced in models with smaller representational capacity, that are more susceptible to overfitting~\cite{blalock2020state}.
Similar trends are echoed in Section~\ref{subsec:additional}.

Consistent results on different ViT models used in our experiments suggest that the proposed approach is not limited to a specific model design, but rather generalizes effectively across distinct architectures and sizes.
This further highlights the utility of the efficiency gains facilitated by sparsifying less informative tokens via \mbox{\textit{SPOT}}, as it can be combined with different ViT models, maximizing resource utilization without sacrificing accuracy.

Additional experimental findings highlighting \mbox{\textit{SPOT}}'s efficacy and robustness, including evaluations of accuracy under image perturbations (Table~\ref{table:perturbations}), attention aggregation strategies comparison (Table~\ref{table:attention_aggregation}), information granularity analysis (Table~\ref{table:granularity_analysis}), smaller-scale model redundant token identification visualizations (Figure~\ref{fig:additional_visualization_T}), and a throughput analysis (Figure~\ref{fig:throughput}), are provided in Appendix~\ref{supp:sec:additional}.

\begin{figure}[t]
    \centering
    \includegraphics[width=.7833\linewidth]{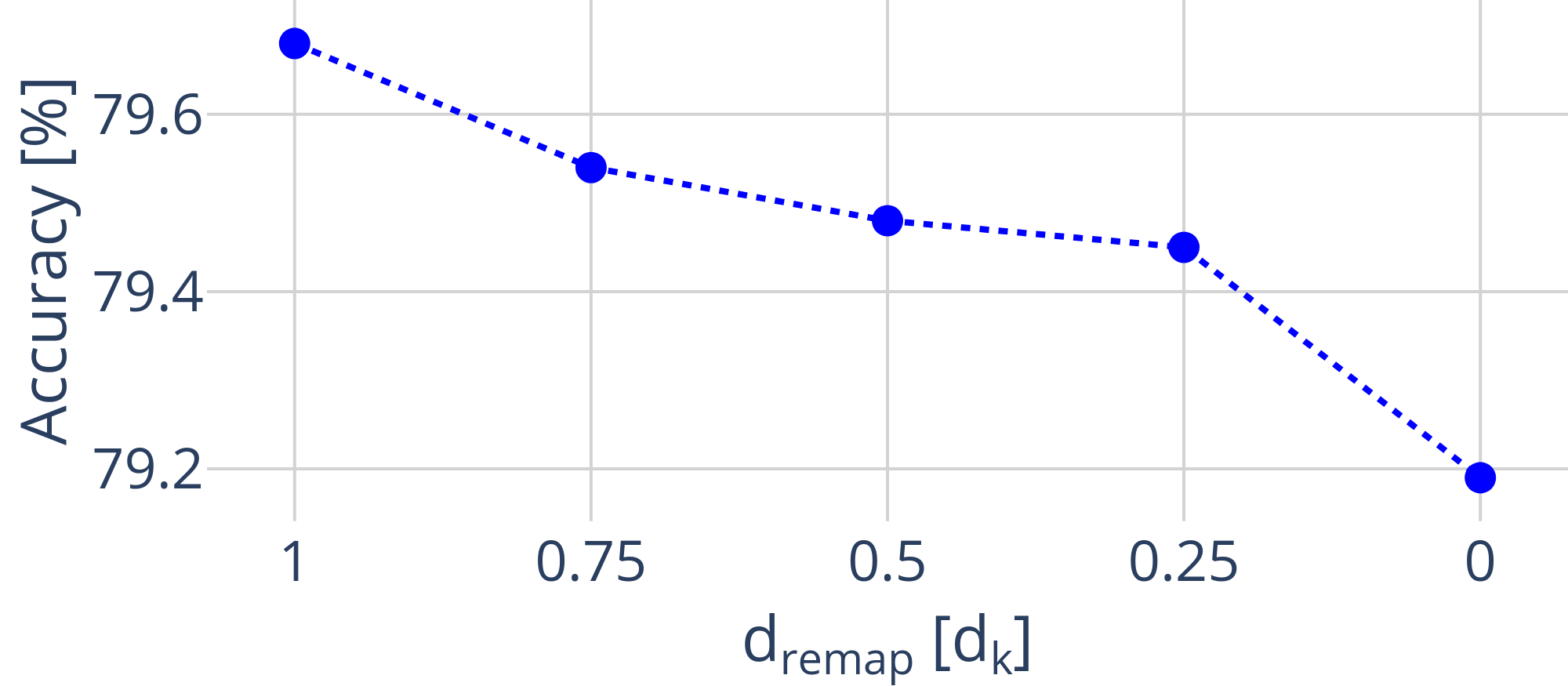}
    \vspace{-0.2cm}
    \caption{Reduced token-derived information results.}
    \label{fig:token_ablation_study}
    \vspace{-0.15cm}
\end{figure}

\subsection{Ablation Study}\label{subsec:ablation_study}
We conduct a set of ablation experiments involving key component modifications and integration with complementary techniques, consistent with the setup in Section~\ref{subsec:setup}.

\begin{table}[b]
    \resizebox{\linewidth}{!}{
    \begin{tabular}{l|lll}
    Model & Method & Accuracy (\%) & GFLOPS \\
    
    \hline
    \multirow{4}{*}{DeiT-T}
     & SPViT*~\cite{kong2022spvit} & 72.0 & 0.8 (-0.5) \\
     & SPViT+\mbox{\textit{SPOT}} (ours) & 72.1 (+0.1) \textcolor{green}{$\blacktriangle$}&  0.8 (-0.5) \\
     & dTPS~\cite{wei2023joint} & 72.9 & 0.8 (-0.5) \\
     & dTPS+\mbox{\textit{SPOT}} (ours) & 73.0 (+0.1) \textcolor{green}{$\blacktriangle$}& 0.8 (-0.5) \\
     
    \hline
    \multirow{4}{*}{DeiT-S}
     & SPViT*~\cite{kong2022spvit} & 79.5 & 2.9 (-1.5) \\
     & SPViT+\mbox{\textit{SPOT}} (ours) & 79.7 (+0.2) \textcolor{green}{$\blacktriangle$} & 2.9 (-1.5) \\
     & dTPS~\cite{wei2023joint} & 80.1 &  3.0 (-1.6)\\
     & dTPS+\mbox{\textit{SPOT}} (ours) &  80.2 (+0.1) \textcolor{green}{$\blacktriangle$}& 3.0 (-1.6)  
    \end{tabular}
    }
    \caption{\mbox{\textit{SPOT}} integration with soft sparsification methods evaluated over ImageNet-1K.
    The observed performance gains underscore the versatility and efficacy of \mbox{\textit{SPOT}} when applied synergistically with complementary techniques.}
    \vspace{-0.06cm}
    \label{table:comparison_soft}
\end{table}

\begin{table*}[t]
    \centering
    \begin{tabular}{l|lllll}
    Model & GFLOPS & CIFAR-100~\cite{krizhevsky2009learning} (\%) & Food-101~\cite{bossard14} (\%) & DTD~\cite{cimpoi2014describing} (\%) & EuroSAT~\cite{helber2018introducing} (\%) \\ \hline
    Baseline~\cite{touvron2021training} & 1.3 & 66.2 & 56.7 & 58.7 & 87.8 \\
    \mbox{\textit{SPOT}} (ours) & 0.8 (-0.5) \textcolor{green}{$\blacktriangledown$} & 66.9 (+0.7) \textcolor{green}{$\blacktriangle$} & 60.7 (+4.0) \textcolor{green}{$\blacktriangle$} & 62.0 (+3.3) \textcolor{green}{$\blacktriangle$} & 90.1 (+2.3) \textcolor{green}{$\blacktriangle$} \\
    \end{tabular}
    \vspace{-0.1cm}
    \caption{Cross-dataset evaluation of \mbox{\textit{SPOT}} using the DeiT-T backbone on diverse datasets under identical fine-tuning settings.
    \mbox{\textit{SPOT}} achieves higher accuracy than the baseline across various visual domains while operating at significantly reduced computational cost.}
    \label{table:cross_datasets}
    \vspace{-0.2cm}
\end{table*}

\noindent\textbf{Reduced Attention-derived Information.}
To understand the impact of each informational element provided to the predictor, we explore the roles of the interaction and dynamics awareness incorporated through the condensed attention information derived from the mean and variance within attention maps.
Calculated across all model layers and attention heads, $\hat{\mu} = \left[ {\mu}^{A, row}, {\mu}^{A, col}, {\mu}^{M, row}, {\mu}^{M, col}, {\mu}^{V, row}, {\mu}^{V, col}\right]$ and $\hat{\sigma} = \left[{{\sigma}^{A, row}}, {{\sigma}^{A, col}}, {{\sigma}^{M, row}}, {{\sigma}^{M, col}}, {{\sigma}^{V, row}}, {{\sigma}^{V, col}}\right]$ represent the means and standard deviations, respectively.
We further investigate the influence of dynamics awareness due to the information from previous layers, $D^M$ and $D^V$.
To this end, we evaluate performance by selectively excluding each of these elements, with the resulting performance variations reported in Table~\ref{table:attention_ablation_study}.
As attention-derived information is heavily condensed, its processing merely changes the computational demands, as reflected by the constant GFLOPS count.
Meanwhile, the consistent decline in accuracy for ablating each element highlights the importance of the holistic approach of \mbox{\textit{SPOT}}, both within each layer and across preceding layers.

\noindent\textbf{Reduced Token-derived Information.}
A key part of the comprehensive dynamics allowing for better-informed detection is token representations, which are projected to a $d_\text{remap}$-dimensional space.
Gradually decreasing $d_\text{remap}$, we reduce the dimensionality of the projected token embeddings.
The GFLOPS count when $d_\text{remap}=0$ is 2.9. Hence, the computational savings from $d_\text{remap}=d_k$ to $d_\text{remap}=0$ (not using token-derived information) are 0.1 GFLOPS.
Results from this analysis (Figure~\ref{fig:token_ablation_study}) indicate that compact attention map representations using our proposed predictor facilitate effective token sparsification, even with limited or no reliance on token embedding information.

\noindent\textbf{Soft Sparsification Integration.}
To further assess the generalizability and modularity of \mbox{\textit{SPOT}}, we integrate it with two common and distinct soft sparsification methods over two model scales, while following their recommended fine-tuning protocols.
To ensure fair comparison, we employ the \mbox{\textit{SPOT}} variant without cross-layer information (corresponding to the setting reported in the third line of Table~\ref{table:attention_ablation_study}), and consistent with the previous experiments, employ sparsification with $\rho = 0.7$.
As summarized in Table~\ref{table:comparison_soft}, \mbox{\textit{SPOT}} consistently improves soft sparsification accuracy while maintaining similar computational budgets.
Although the gains may appear modest in absolute terms, they are achieved in settings where existing methods already saturate performance and leave little room for improvement, underscoring \mbox{\textit{SPOT}}'s effectiveness in soft sparsification regimes as well.

\subsection{Comparison over Additional Benchmarks}\label{subsec:additional}
To further validate the versatility and transferability of \mbox{\textit{SPOT}}, we conducted evaluations across diverse image datasets, with their key properties summarized in Table~\ref{table:datasets}.
These datasets collectively cover a broad range of domains, allowing assessment under varying visual semantics and data distributions.

For this analysis, we compare the baseline DeiT-T model initialized from its publicly available ImageNet-1K pre-trained weights, and the \mbox{\textit{SPOT}}-augmented DeiT-T model corresponding to the configurations reported in Table~\ref{table:comparison_hard}.
Full training details are provided in Appendix~\ref{supp:subsubsec:dataset_ablation}.

Importantly, the objective of this experiment is to assess the adaptability and cross-domain generalization capacity of \mbox{\textit{SPOT}} compared to the baseline under identical settings.
Thus, it provides a fair evaluation of its transfer behavior, rather than maximizing dataset-specific performance.

As summarized in Table~\ref{table:cross_datasets}, \mbox{\textit{SPOT}} consistently exceeds baseline accuracy across diverse datasets while operating at significantly lower computational costs.
Under identical fine-tuning settings, \mbox{\textit{SPOT}}-augmented models adapt faster to distributional shifts and new visual domains in smaller datasets.
Its conditioning on both intra-layer and cross-layer attention dynamics, combined with token-level information, allows it to preserve semantically informative regions that generalize across diverse domains, while discarding redundant tokens\hide{ in a context-aware fashion}, as evidenced by overall performance.

These findings suggest that \mbox{\textit{SPOT}} not only maintains strong performance on large-scale datasets such as ImageNet, but also transfers effectively to smaller or domain-specific datasets, highlighting its robustness and suitability as a well-generalized token sparsification framework.

\section{Conclusions}\label{sec:conclusions}
We presented an effective and robust framework for token importance prediction, designed to improve the efficiency of Vision Transformers (ViT).
The proposed \textbf{SP}arsification with attenti\textbf{O}n dynamics via \textbf{T}oken relevance (\mbox{\textit{SPOT}}) framework leverages both token and attention-derived dynamics information propagated across the layers of the ViT architecture.
This mechanism enables to \mbox{\textit{SPOT}} salient tokens while pruning contextually redundant ones, yielding sparsification patterns that align with human-interpretable feature relevance.
Through comprehensive empirical experiments conducted using two ViT-based architectures with different model sizes, evaluations over diverse visual domain benchmarks, and integration with complementary efficiency enhancement methods, \mbox{\textit{SPOT}} outperforms existing methods in ViT sparsification informed via salient token identification.
Critically, \mbox{\textit{SPOT}} achieves a substantial reduction in computational cost, up to 40\%, while maintaining near-identical or even improving model accuracy.


{
    \small
    \bibliographystyle{ieeenat_fullname}
    \bibliography{main}
}

\clearpage
\setcounter{page}{1}
\maketitlesupplementary

\section{Token Relevance Module}\label{supp:sec:module}
\subsection{Architecture}\label{supp:subsec:architecture}
The token identification module first employs the following layers to generate 
$z_\text{local}$ and $z_\text{global}$ token representations:
\[
\text{LayerNorm} \;\to\; \text{Linear}(d_\text{remap}, \tfrac{d_\text{remap}}{2}) \;\to\; \text{GELU},
\]
where GELU refers to a Gaussian Error Linear Unit~\cite{hendrycks2016gaussian}.

In addition to these base features, it incorporates attention-derived statistics from the attention maps in relevant heads and layers, $D_{l,h}^A$.
The token and attention components are then concatenated into a unified representation.

Finally, the module applies
\[
\begin{aligned}
&\text{Linear}(E, \tfrac{E}{2}) \;\to\; \text{GELU} 
   \;\to\; \text{Linear}(\tfrac{E}{2}, \tfrac{E}{4}) \;\to\; \text{GELU} \\
&\to\; \text{Linear}(\tfrac{E}{4}, 2) \;\to\; \text{Softmax},
\end{aligned}
\]
which transforms this concatenated vector into a probability distribution from which redundant tokens are identified.  

The embedding dimension $E$ depends on the inclusion of attention-derived features, as detailed in Sections~\ref{subsec:method} and~\ref{subsec:ablation_study}.
The design further allows incorporating statistics from earlier processing stages, which may include mean-based descriptors, variance-based descriptors, or both.
Similarly, these statistics can be computed separately for each head or once across all heads, with the latter reducing the resulting feature dimensionality.

A key distinction of our module design lies in its capacity to process and utilize compact attention map representations for token relevance prediction.
This approach enables more effective identification of contextually redundant tokens and delivers notable performance improvements compared to sparsification techniques that depend solely on token embedding or attention-derived information.

\subsection{Implementation Details}\label{supp:subsec:implementation}
\subsubsection{General Experimental Setup}\label{supp:subsubsec:general}
The following implementation details apply to all experiments unless otherwise specified.

All models were optimized using AdamW~\cite{loshchilov2018decoupled}.
The network was fine-tuned using eight NVIDIA GeForce RTX 4090 GPUs with a learning rate of 0.001, while the prediction modules used a rate of 0.01 to facilitate enhanced adaptation.
Unless otherwise stated, all model instances underwent fine-tuning without a distillation token for 80 epochs with a batch size of 512.
When integrating our approach with other methods, training durations were adjusted to align with their corresponding recommended fine-tuning protocols for fair comparison.
For all experiments, we used a patch size of $16 \times 16$ pixels for processing.
For the ImageNet~\cite{deng2009imagenet} experiments, the input image resolution was $224 \times 224$ pixels.

While DeiT~\cite{touvron2021training} preserves a structure comparable to standard ViTs, LV-ViT~\cite{jiang2021all} introduces token labeling as an auxiliary training objective and incorporates convolutional operations into its design, requiring similarity between the refined tokens and those of a pre-trained model.
Hence, the loss coefficients $\lambda_1$ and $\lambda_2$ in Equation~\ref{eq:objective_function} were set to 2 and 0.5, respectively, across all experiments, while $\lambda_3$ was set to 0 with DeiT models and 10 with LV-ViT models.
The dimensionality of remapped token representations, $d_\text{remap}$, was set to the original dimensionality of the token embedding space, $d_{k}$, for each respective model.

In accordance with established practices in prior literature~\cite{rao2021dynamicvit, kong2022spvit, wei2023joint}, token relevance prediction and subsequent sparsification were applied one layer after each quarter mark of the network depth, for the first three quarters, resulting in $K=3$ sparsification stages.
Specifically, for models with 12 layers, sparsification modules were plugged into layers 4, 7, and 10, whereas for the 16-layer LV-ViT-S, they were analogously positioned at layers 5, 9, and 13.
Each sparsification iteration employs a separate module, as detailed in Section~\ref{subsec:method}, with a token retention rate $\rho_k$.

\begin{table}[t]
    \resizebox{\linewidth}{!}{
    \begin{tabular}{l|lll}
    Model & Method & Accuracy (\%) & GFLOPS \\
    
    \hline
    \multirow{2}{*}{DeiT-T}
     & Baseline~\cite{touvron2021training} & 44.8 & 1.3 \\
     & \mbox{\textit{SPOT}} (ours) & 45.2 (+0.4) \textcolor{green}{$\blacktriangle$} & 0.8 (-0.5) \textcolor{green}{$\blacktriangledown$} \\
     
    \hline
    \multirow{2}{*}{DeiT-S}
     & Baseline~\cite{touvron2021training} & 57.8 & 4.6 \\
     & \mbox{\textit{SPOT}} (ours) & 58.0 (+0.2) \textcolor{green}{$\blacktriangle$} & 3.0 (-1.6) \textcolor{green}{$\blacktriangledown$} \\

    \hline
    \multirow{2}{*}{LV-ViT-T}
     & Baseline~\cite{jiang2021all} & 50.8 & 2.9 \\
     & \mbox{\textit{SPOT}} (ours) & 49.6 (-1.2) \textcolor{red}{$\blacktriangledown$} & 2.0 (-0.9) \textcolor{green}{$\blacktriangledown$} \\

    \hline
    \multirow{2}{*}{LV-ViT-S}
     & Baseline~\cite{jiang2021all} & 60.3 & 6.6 \\
     & \mbox{\textit{SPOT}} (ours) & 61.0 (+0.7) \textcolor{green}{$\blacktriangle$} & 4.5 (-2.1) \textcolor{green}{$\blacktriangledown$} \\
    \end{tabular}
    }
    \vspace{-0.1cm}
    \caption{Results of robustness evaluation on DeiT and LV-ViT models over ImageNet-C.
    \mbox{\textit{SPOT}} not only achieves substantial computational savings but also improves accuracy in most cases compared to the baselines, demonstrating both efficiency and stability under perturbations.}
    \vspace{-0.2cm}
    \label{table:perturbations}
\end{table}

\subsubsection{Cross-Domain Ablation Setup}\label{supp:subsubsec:dataset_ablation}
For the ablation study concerning additional benchmarks in Section~\ref{subsec:ablation_study}, the following specific fine-tuning protocol was used.
For each dataset, each model classification head was fine-tuned for 80 epochs with a learning rate of 0.025, while freezing the remaining components.
This fine-tuning configuration is designed to encourage adaptation on smaller-scale datasets rather than subtle adjustments of pre-trained weights.
No teacher supervision is applied in these downstream tasks, as the teacher model was trained solely on ImageNet; its classification head is specifically configured for ImageNet class labels and could bias adaptation dynamics.
Although the DeiT models used in this setup were not trained exactly following the original recipe~\cite{touvron2021training}, both the baseline and \mbox{\textit{SPOT}}-augmented models share identical architectural and training conditions.
This ensures that performance differences reflect \mbox{\textit{SPOT}}’s true effect.

\section{Additional Experiments}\label{supp:sec:additional}
In this section, we aim to rigorously isolate and characterize the intrinsic robustness behavior of the proposed \mbox{\textit{SPOT}} framework, independent of architectural or training modifications introduced by other methods.
Unlike the comparisons against state-of-the-art sparsification and token selection approaches presented in the main manuscript, the following evaluations focus on \mbox{\textit{SPOT}}’s behavior relative to standard, well-established baselines.
While prior works on token sparsification have primarily emphasized performance on ImageNet-1k, aspects such as robustness to perturbations and cross-domain transferability have received comparatively limited attention.
In contrast, our evaluations explicitly address these aspects, enabling observed differences in robustness, accuracy, and computational efficiency to be directly attributed to \mbox{\textit{SPOT}} itself, whose design facilitates the early detection and suppression of less informative or contextually redundant tokens.
This setup provides a controlled and reproducible assessment that serves as a complementary and orthogonal analysis to the state-of-the-art comparisons presented in the main manuscript.

\begin{table}[t]
    \centering
    \resizebox{\linewidth}{!}{
    \begin{tabular}{l|llll}
    Dataset & \makecell{Training\\Images} & \makecell{Test\\Images} & Categories & Domain \\

    \hline

    ImageNet-1K~\cite{deng2009imagenet} & 1{,}281{,}167 & 50{,}000 & 1{,}000 & Natural images \\
    ImageNet-C~\cite{hendrycks2019benchmarking} & -- & 950{,}000 & 1{,}000 & Corrupted images \\
    CIFAR-100~\cite{krizhevsky2009learning} & 50{,}000 & 10{,}000 & 100 & Pixelated objects \\
    Food-101~\cite{bossard14} & 75{,}750 & 25{,}250 & 101 & Food items \\
    DTD~\cite{cimpoi2014describing} & 3{,}760 & 1{,}880 & 47 & Textures \\
    EuroSAT~\cite{helber2018introducing} & 21{,}600 & 5{,}400 & 10 & Satellite images \\
\end{tabular}
}
\vspace{-0.1cm}
\caption{Summary of datasets used for evaluation. The listed numbers reflect our usage of these datasets in this work. \mbox{ImageNet-C} was used solely for evaluation.}
\label{table:datasets}
\vspace{-0.2cm}
\end{table}

\subsection{Robustness under Perturbations}
To further assess the transferability and stability of \mbox{\textit{SPOT}}, we evaluate its performance under challenging, corrupted visual conditions using the ImageNet-C benchmark~\cite{hendrycks2019benchmarking}.
This dataset consists of ImageNet-1k dataset images, which have been independently subjected to a diverse suite of 19 image degradation types, including noise, blur, weather, and digital distortions, applied at varying severity levels (1-5).

We adopt an intermediate severity level (3), which provides a balanced configuration that introduces meaningful degradations while preserving image recognizability.
This setting enables a realistic assessment of robustness without saturating error rates or obscuring relative performance trends.
All evaluations were conducted while employing the \mbox{\textit{SPOT}} models fine-tuned on ImageNet-1K, as detailed in Section~\ref{sec:experiments} and reported in Table~\ref{table:comparison_hard}, using the same computational budgets and without any additional adaptation or fine-tuning.
Model accuracy is reported in Table~\ref{table:perturbations} to quantify stability across perturbations relative to the baseline.
Unless otherwise specified, all reported GFLOPS values denote total computational cost (equivalent to ''Overall GFLOPS'').

Across models, \mbox{\textit{SPOT}} consistently maintains comparable accuracy to the baseline, and in most cases surpasses it, demonstrating strong robustness to input noise and distributional shifts despite operating at significantly lower computational budgets.
This behavior can be attributed to \mbox{\textit{SPOT}}’s estimation and utilization of most relevant tokens, which can better preserve the signal in the input despite perturbations, mitigating the amplification of corrupted or spurious tokens at a certain layer.
Consequently, when exposed to input perturbations, the model’s predictions are guided by a more robust and context-aware representation of token dynamics, which better preserves the integrity of salient feature representations and yields higher accuracy despite reduced computational capacity.
These mechanisms underpin \mbox{\textit{SPOT}}’s stable and broadly applicable computational benefits.

The results indicate that \mbox{\textit{SPOT}} generalizes effectively beyond clean visual conditions, exhibiting resilience to both high-frequency corruptions (e.g., impulse noise) and low-frequency distortions (e.g., fog).
This robustness underscores that \mbox{\textit{SPOT}}’s attention dynamics regularization implicitly contributes to robust modeling of salient dependencies between tokens, reinforcing its suitability for deployment in real-world, noisy visual environments.

\subsection{Information Aggregation and Granularity Analysis}
We analyze the impact of different information aggregation strategies, comparing our learnable predictor against a non-learnable heuristic and evaluating the effect of information granularity within the learnable model.

First, to isolate the contribution of \mbox{\textit{SPOT}}’s learnable attention-derived information-based importance prediction from simpler aggregation, we implemented a deterministic, non-learnable variant of \mbox{\textit{SPOT}}, comparing both under identical conditions on the DeiT-S backbone and without token-derived information.

\begin{table}[t]
    \resizebox{\linewidth}{!}{%
    \centering
    \begin{tabular}{l|lll}
    Attention aggregation & \begin{tabular}[c]{@{}l@{}}Accuracy\\ (\%)\end{tabular} & \begin{tabular}[c]{@{}l@{}}Predictor\\ GFLOPS\end{tabular} & \begin{tabular}[c]{@{}l@{}}Overall\\ GFLOPS\end{tabular} \\ \hline
    Baseline~\cite{touvron2021training} & 79.8 & -- & 4.6 \\
    Averaging across layers & 73.6 (-6.2) & -- & 2.9 (-1.7) \\
    \textbf{\mbox{\textit{SPOT}}} (ours) & \textbf{79.2 (-0.6)} & \textless0.01 & 2.9 (-1.7)
    \end{tabular}
    }
    \caption{Comparison between learnable and non-learnable (heuristic) attention aggregation strategies on the DeiT-S backbone.
    ‘--’ on the Predictor GFLOPS column denotes that no learnable predictor is used.
    }
    \label{table:attention_aggregation}
    \vspace{-0.2cm}
\end{table}

In this heuristic configuration, the learnable predictor was replaced by a fixed, attention-averaging mechanism, thereby eliminating contributions from learned weighting of attention-derived features.
At each sparsification stage, token importance was computed as the mean class-token attention score ($A_{l,h}^{cls, out}$), averaged across all attention heads and all preceding layers ($l = 1, \ldots, k$).
This scalar value captures the cumulative amount of attention a token has received from the class token throughout the network up to stage $k$, integrating attention information over multiple layers and heads.
The $\rho_k$ tokens exhibiting the highest mean scores were deterministically retained, while the remaining tokens were pruned, in accordance with attention-based token redundancy reduction principles, as detailed in Section~\ref{sec:related_work}.
To ensure a fair comparison, the DeiT-S backbone was fine-tuned following the same optimization setup described in Section~\ref{subsec:setup} and Appendix~\ref{supp:subsec:implementation}.

As summarized in Table~\ref{table:attention_aggregation}, the described heuristic-based approach achieved an accuracy of 73.6\% with a computational cost of 2.9 GFLOPS when evaluated on the ImageNet-1K dataset.
This reflects a notable drop in accuracy relative to the baseline and to \mbox{\textit{SPOT}} ablation variant that retained the learnable predictor while similarly excluding token-derived information, as detailed in the reduced token-derived information ablation in Section~\ref{subsec:ablation_study} for $d_\text{remap}=0$.
The mentioned ablation attained an accuracy of 79.2\%, as illustrated in Figure~\ref{fig:token_ablation_study}.
Importantly, this degradation in accuracy does not correspond to a noticeable computational benefit, since both configurations exhibit an equivalent computational cost of 2.9 GFLOPS (as reported in Section~\ref{subsec:ablation_study}), given that the processing of attention-derived features within the proposed predictor incurs only negligible computational overhead.

This finding demonstrates that \mbox{\textit{SPOT}}'s efficacy cannot be attributed solely to multi-layer attention aggregation, but rather to its learnable modeling of attention dynamics.
This ablation confirms that the \mbox{\textit{SPOT}} predictor captures complementary, context-aware information not fully perceptible to fixed, non-learnable aggregation heuristics, thereby reinforcing the validity of our proposed learnable sparsification framework.

We note that this variant is inspired by earlier work in token pruning for vision transformers that uses class‐token attention to score and prune tokens in a single‐layer fashion, e.g., EViT~\cite{liang2022not}.

Second, we explore the efficacy of incorporating more granular information and modules versus aggregated ones.
The results for granularity experiments are summarized in Table~\ref{table:granularity_analysis}.

The default \mbox{\textit{SPOT}} configuration feeds per-head attention values to its predictor.
In a multi-head attention layer with $H$ heads, each head learns different types of token relationships.
By providing all $H$ sets of attention-derived features, the predictor has access to this rich, diverse information.
To this end, aiming to check how providing averaged attention values across the multi-head attention $H$ heads, $D_{l} = \frac{1}{H} \cdot \sum_{h=1}^{H}D_{l,h}$, performs compared to per-head attention values.

Finally, we investigate the importance of an iteration-specific learnable token relevance prediction module.
The default \mbox{\textit{SPOT}} configuration includes separate predictor modules for each stage.
For instance, the predictor at the first sparsification stage is trained specifically to find unimportant tokens based on information up to its corresponding layer, and does not share weights with other learnable modules.
This experiment employs one common token detection module.
This single, shared predictor module with one set of weights is being trained and applied repeatedly at all $K$ sparsification stages.

\begin{table}[t]
    \resizebox{\linewidth}{!}{%
    \centering
    \begin{tabular}{l|lll}
    Predictor modification & \begin{tabular}[c]{@{}l@{}}Accuracy\\ (\%)\end{tabular} & \begin{tabular}[c]{@{}l@{}}Predictor\\ GFLOPS\end{tabular} & \begin{tabular}[c]{@{}l@{}}Overall\\ GFLOPS\end{tabular} \\ \hline
    Head-averaged attention & 79.5 (-0.3) & 0.106 & 3.0 (-1.6) \\
    Iterations-shared module & 79.4 (-0.4) & 0.129 & 3.0 (-1.6)
    \end{tabular}
    }
    \caption{Granularity analysis results of \mbox{\textit{SPOT}} on the DeiT-S backbone contrasting two module modifications: using head-averaged attention values versus per-head inputs, and using a single, shared predictor module versus stage-specific modules.    
    }
    \label{table:granularity_analysis}
    \vspace{-0.2cm}
\end{table}

\subsection{Qualitative Results}
\begin{figure*}[t]
    \centering
    \begin{tabular}{cc}
        \vspace{-0.05cm}
        \includegraphics[width=0.46\linewidth]{figures/visualization_S/Text2.png} & \hspace{-0.4cm} \includegraphics[width=0.46\linewidth]{figures/visualization_S/Text2.png} \\
        \vspace{-0.05cm}
        \includegraphics[width=0.46\linewidth]{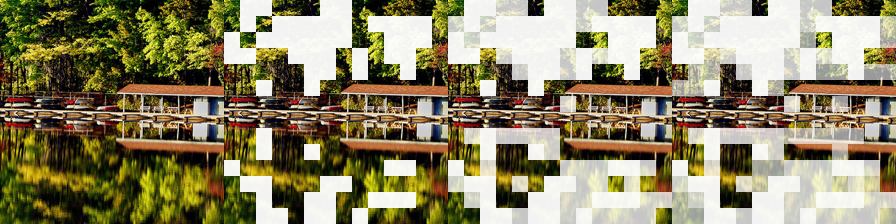} & \hspace{-0.4cm} \includegraphics[width=0.46\linewidth]{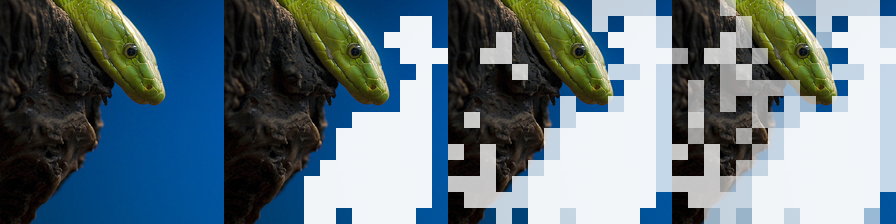} \\
        \vspace{-0.05cm}
        \includegraphics[width=0.46\linewidth]{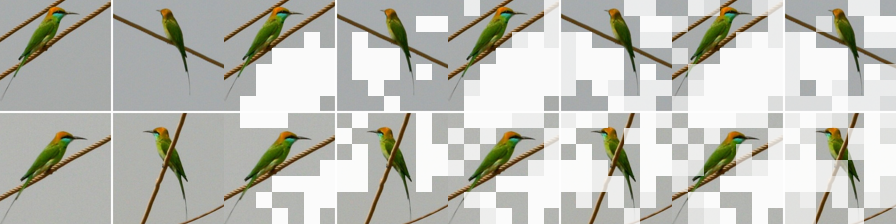} & \hspace{-0.4cm} \includegraphics[width=0.46\linewidth]{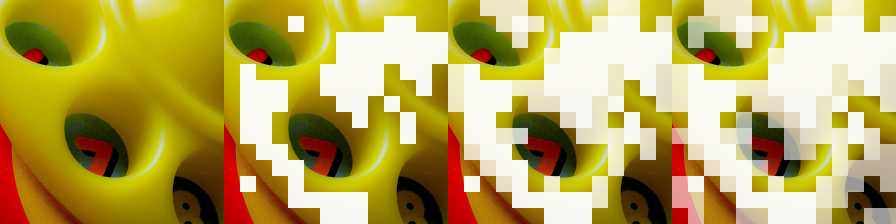} \\
        \vspace{-0.05cm}
        \includegraphics[width=0.46\linewidth]{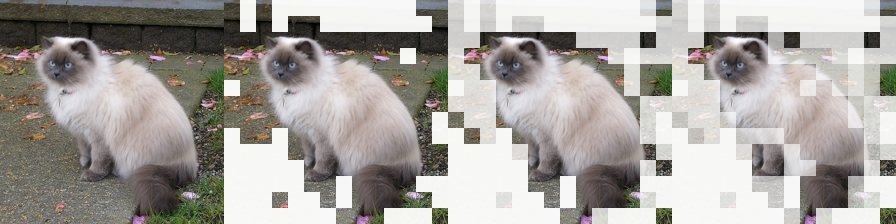} & \hspace{-0.4cm} \includegraphics[width=0.46\linewidth]{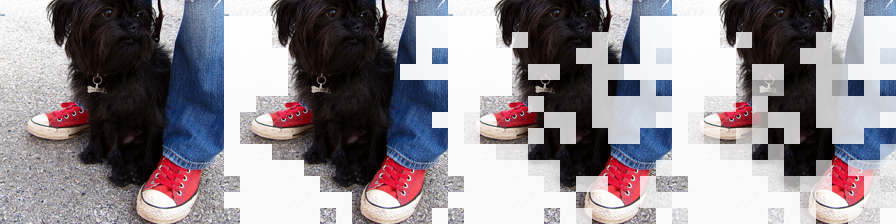} \\
        \vspace{-0.05cm}
        \includegraphics[width=0.46\linewidth]{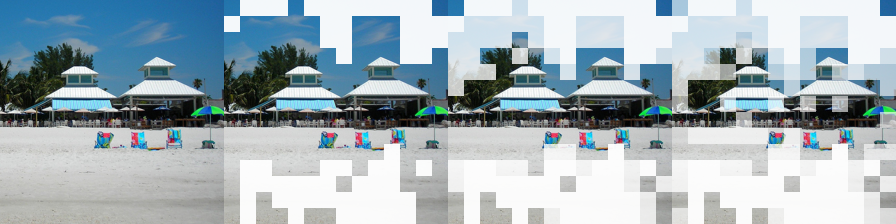} & \hspace{-0.4cm} \includegraphics[width=0.46\linewidth]{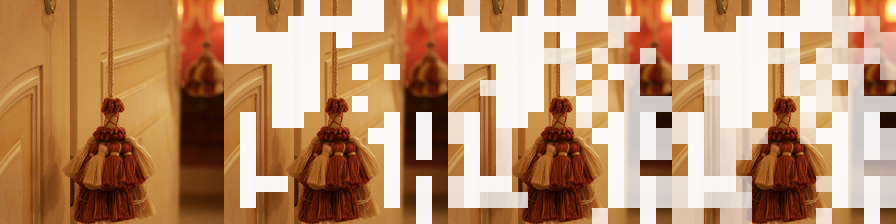} \\
        \vspace{-0.05cm}
        \includegraphics[width=0.46\linewidth]{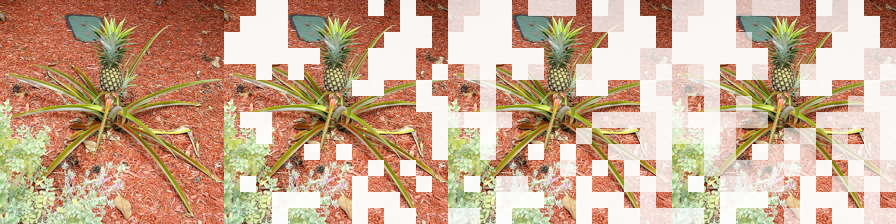} & \hspace{-0.4cm} \includegraphics[width=0.46\linewidth]{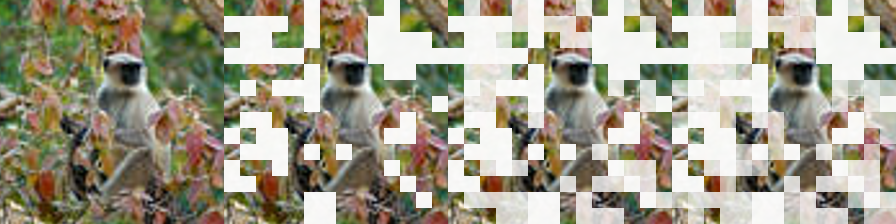} \\
        \vspace{-0.05cm}
        \includegraphics[width=0.46\linewidth]{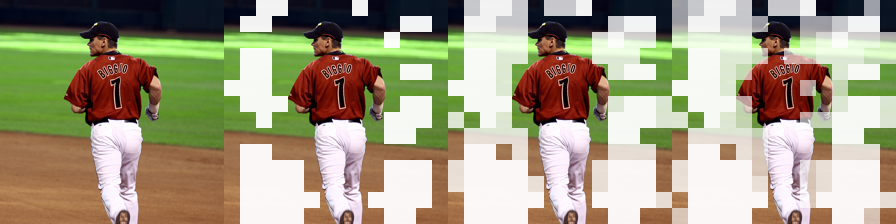} & \hspace{-0.4cm} \includegraphics[width=0.46\linewidth]{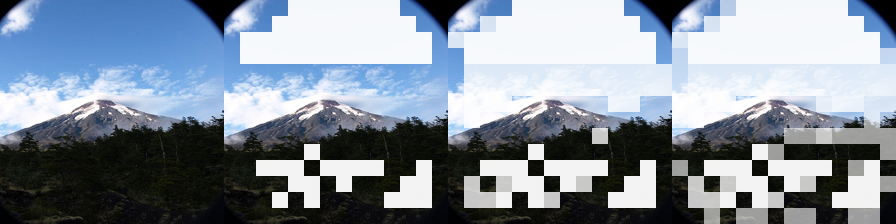} \\
    \end{tabular}
    \vspace{-0.15cm}
    \caption{Visualizations of the gradual redundant token detection performed by our proposed approach on DeiT-T on samples from ImageNet-1K validation set.
    Increasingly transparent masking shades indicate later detection.
    Tokens identified as more informative, and thereby retained, are well aligned with semantic image objects and visual features, pointing to \mbox{\textit{SPOT}}'s interpretability.}
    \label{fig:additional_visualization_T}
\end{figure*}

Visualizations of the token identification process applied to samples from the ImageNet-1K dataset for the DeiT-T model are presented in Figure~\ref{fig:additional_visualization_T}.
The integration of \mbox{\textit{SPOT}} with the DeiT-T architecture demonstrably preserves class-discriminative tokens despite its smaller capacity compared to the DeiT-S model, whose results are detailed in Figure~\ref{fig:visualization}.
This observation further underscores the interpretability and generalization capabilities of the proposed methodology.

\subsection{Throughput Analysis}
\begin{figure*}[t]
    \centering
    \includegraphics[width=0.85\textwidth]{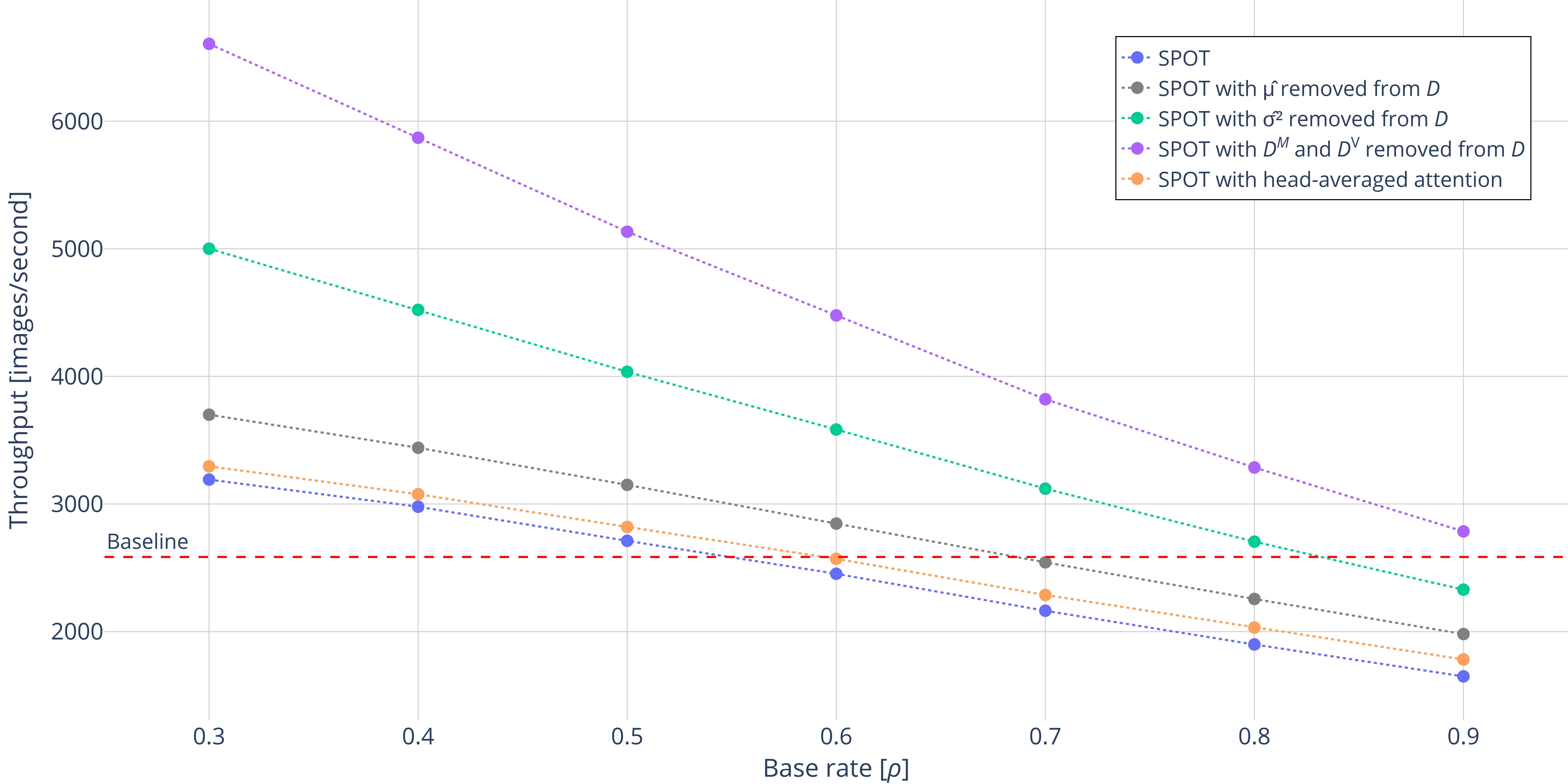}
    \caption{Comparative throughput analysis of \mbox{\textit{SPOT}} integrated with the DeiT-S architecture, comparing different variants of the proposed \mbox{\textit{SPOT}} redundant token prediction module.
    The evaluation was performed on a single NVIDIA GeForce RTX 4090 GPU, employing a batch size of 512.
    Each variant is characterized by the distinct combination of token and attention-derived information it incorporates for prediction.
    This design ensures that \mbox{\textit{SPOT}} is adaptable, allowing users to calibrate its parameters to meet specific computational constraints and performance objectives.}
    \label{fig:throughput}
\end{figure*}

Furthermore, we present an empirical throughput analysis for the various \mbox{\textit{SPOT}} variants, which are characterized by the quantity and nature of the information received as input.
Consistent with the variants examined in Section~\ref{subsec:ablation_study}, Figure~\ref{fig:throughput} provides a comprehensive overview of the empirical throughput performance of \mbox{\textit{SPOT}} when integrated with the DeiT-S architecture with different token retention rates, denoted as $\rho$.
All measurements were obtained under the experimental configuration outlined in Section~\ref{subsec:setup}.
It is important to note that the throughput of \mbox{\textit{SPOT}} is largely dictated by the underlying base architecture or method with which it is integrated, since \mbox{\textit{SPOT}} does not modify the architecture or the computational pipeline.
Accordingly, we focus our throughput analysis on comparing different \mbox{\textit{SPOT}} variants within the same architecture (DeiT-S), thereby isolating the impact of its redundant-token prediction mechanisms.

As illustrated in Figure~\ref{fig:throughput}, \mbox{\textit{SPOT}}'s versatility allows for a range of throughput levels that can be calibrated to meet different computational constraints and performance objectives.
Such objectives are characterized by the token retention rate, $\rho$, and the predictor’s input composition, as empirically validated in Section~\ref{sec:experiments}.
By selectively incorporating or removing token and attention-derived information, \mbox{\textit{SPOT}} enables a trade-off between throughput and accuracy across varying retention rates.
For instance, one \mbox{\textit{SPOT}} variant achieves a throughput of 3,880 images per second, over x1.5 higher compared to the baseline, while operating at 3.0 GFLOPS and maintaining an accuracy of 79.5\%.
This demonstrates the flexibility of \mbox{\textit{SPOT}} in adapting to diverse operational settings while preserving strong performance.

\section{Theoretical Perspective}\label{supp:sec:theoretical}
The efficacy of \mbox{\textit{SPOT}} can be interpreted through the lens of statistical reduction of sample variance and structured aggregation of attention signals.

Guo et al.~\cite{guo2023robustifying} show that in Vision Transformers (ViTs), a single attention layer provides an inherently noisy estimation of a token's underlying importance, denoted as $R[i]$, and that ViTs often suffer from unstable distributions of high-attention tokens, where even mild perturbations can significantly alter which tokens appear salient.
The attention scores $A_l[i]$ assigned to token $i$ at layer $l$ within a specific model head, and by other tokens, can be modeled as
$$
A_{l}[i] = R[i] + \epsilon_{l}[i],
$$
where $\epsilon_{l}[i]$ represents fluctuations arising from spurious spikes or localized instabilities.

When token relevance is inferred from one layer, as common in prior literature, the variance of the estimator is $\mathrm{Var}(A_{l}[i]) = \sigma_{\epsilon}^{2}$.
Conversely, \mbox{\textit{SPOT}} aggregates across $L$ layers to formulate a composite estimator:
$$
\hat{R}[i] = \frac{1}{L} \sum_{l=1}^{L} A_{l}[i].
$$
This aggregation strategy substantially reduces the estimator's variance, which becomes:
$$
\mathrm{Var}(\hat{R}[i]) =
\mathrm{Var}\left(\frac{1}{L} \sum_{l=1}^{L} A_{l}[i]\right) \approx
\frac{1}{L^2} L \sigma_{\epsilon}^{2}=
\frac{\sigma_{\epsilon}^{2}}{L},
$$
under the assumption of weakly correlated noise across different layers.
Thus, inter-layer aggregation functions as a form of ensemble averaging~\cite{hashem1997optimal}, yielding a more robust and lower-variance estimate of token saliency.
Importantly, this procedure is performed independently for each attention head, after which the resulting descriptors are further aggregated across all heads, ensuring that complementary interaction patterns captured by distinct heads are synergistically leveraged.

As detailed in Section~\ref{sec:methods}, \mbox{\textit{SPOT}} compresses attention maps into compact descriptors that capture both outgoing and incoming dynamics.
These descriptors are constructed from several components, including the first row and column of the attention map, which quantify the distribution of outgoing influence from the global aggregator, and the contribution of each patch token to the class token, respectively.
To further summarize interaction patterns, the framework computes row-wise and column-wise statistics, namely the mean and variance (formulated in equations~\ref{eq:mu_row}, ~\ref{eq:mu_col}, ~\ref{eq:sigma_row}, ~\ref{eq:sigma_col}).
The row-wise moments encode how a given token distributes its attention across all other tokens, while the column-wise moments describe how a token is attended to by others, thereby capturing its global significance.

The utilization of these moment-based descriptors yields two principal advantages.
First, by summarizing attention distributions with statistical moments, the influence of spurious spikes and outliers within the raw attention maps is suppressed.
Second, the explicit differentiation between row-wise and column-wise patterns provides crucial directional context, enabling \mbox{\textit{SPOT}} to distinguish between tokens that broadly influence others and those that consistently attract attention.

Beyond single-layer descriptors, \mbox{\textit{SPOT}} also models the trajectories of these statistical moments across layers.
The mean of these statistics across layers, $D_{l,h}^M$, serves as an indicator of a token’s influence and cumulative saliency throughout the model, highlighting which tokens consistently exert meaningful impact on information flow.
Simultaneously, the variance of these values across layers, $D_{l,h}^\Sigma$, measures the stability of a token's role, indicating whether its importance is consistent or fluctuates with local context.
This characterization of inter-layer attention evolution helps prevent the premature pruning of tokens that may become relevant in deeper layers of the network.

By combining the raw first row and column vectors, summary statistics from row and column-wise moments, aggregated cross-layer moments, and token embeddings from the current layer, \mbox{\textit{SPOT}} constructs a compact yet information-rich representation of tokens and their attention dynamics.
This composite representation comprehensively encodes both local and global interactions, tracking short-term fluctuations and long-term stability, while remaining resilient to transient attention noise.
Thus, \mbox{\textit{SPOT}} offers a holistic and theoretically-supported framework for more interpretable token sparsification.

\end{document}